\documentclass[final,5p,times,twocolumn]{elsarticle}

\usepackage{amssymb}
\usepackage{multirow}
\usepackage{caption} 	 			
\usepackage{booktabs} 		
\usepackage{textcomp}
\usepackage{hyperref}
\usepackage{amsmath,amsfonts}
\usepackage{algorithmic}
\usepackage{algorithm}
\usepackage{array}
\usepackage{hhline}
\usepackage{fancyhdr}

\journal{arxiv}
\begin{document}

\begin{frontmatter}



\title{MS$^{2}$3D: A 3D Object Detection Method Using Multi-Scale Semantic Feature Points to Construct 3D Feature Layer}


\author[label1]{Yongxin Shao}
\author{Aihong Tan\corref{cor2}\fnref{label1}}
\author[label1]{Binrui Wang}
\author{Tianhong Yan\corref{cor1}\fnref{label1}}
\author[label1]{Zhetao Sun}
\author[label1]{Yiyang Zhang}
\author[label1]{Jiaxin Liu}
\affiliation[label1]{organization={The School of Mechanical and Electrical Engineering, China Jiliang Universty},
            city={Hanzhou},
            country={China}}
\cortext[cor1]{Secondary Corresponding author, E-mail address : thyan@163.com (Tianhong Yan).}
\cortext[cor2]{Principal corresponding author, E-mail address : Tanah@cjlu.edu.cn (Aihong Tan).}

\begin{abstract}
LiDAR point clouds can effectively depict the motion and posture of objects in three-dimensional space. Many studies accomplish the 3D object detection by voxelizing point clouds. However, in autonomous driving scenarios, the sparsity and hollowness of point clouds create some difficulties for voxel-based methods. The sparsity of point clouds makes it challenging to describe the geometric features of objects. The hollowness of point clouds poses difficulties for the aggregation of 3D features. We propose a two-stage 3D object detection framework, called MS$^{2}$3D. (1) We propose a method using voxel feature points from multi-branch to construct the 3D feature layer. Using voxel feature points from different branches, we construct a relatively compact 3D feature layer with rich semantic features. Additionally, we propose a distance-weighted sampling method, reducing the loss of foreground points caused by downsampling and allowing the 3D feature layer to retain more foreground points. (2) In response to the hollowness of point clouds, we predict the offsets between deep-level feature points and the object's centroid, making them as close as possible to the object's centroid. This enables the aggregation of these feature points with abundant semantic features. For feature points from shallow-level, we retain them on the object's surface to describe the geometric features of the object. To validate our approach, we evaluated its effectiveness on both the KITTI and ONCE datasets.
\end{abstract}

%

\begin{keyword}
3D object detection \sep Point clouds \sep LiDAR \sep Deep learning \sep



\end{keyword}

\end{frontmatter}


\section{Introduction}

\thispagestyle{fancy}
\renewcommand{\headrulewidth}{0pt}
\lhead{\centering{\small{Copyright \copyright 2024 Elsevier. This article has been accepted for publication in a future issue of this journal, but has not been fully edited. Content may change prior to final publication. Citation information: DOI https://doi.org/10.1016/j.neunet.2024.106623 , Neural Networks.}}}

With the continuous development of environmental perception technology \cite{bochkovskiy2020yolov4,SANTHAKUMAR2022167,MENEZES2023476,AHISHALI202315, s22031098, masood2022recognition} and 3D sensors \cite{sun2020disp,chen2020dsgn,chang2018pyramid,li2019stereo, ZOU2023609, MUKHTAR2023363}, the application of LiDAR in autonomous driving is gradually maturing. The point clouds generated by LiDAR can effectively describe the geometric shape and motion pose of objects in three-dimensional space \cite{chen20153d,deng2017amodal,xu2017learning,gupta2014learning,shao2023efficient,9773165}. However, point clouds exhibit distinct nature \cite{qi2017pointnet, YAO2023350, LIU2024263, rs16020327}, such as sparsity, disorder, and rotation invariance, pose challenges for 3D object detection in autonomous driving scenarios.

\begin{figure}[!t]
	\centering
	\includegraphics[width=3.5in]{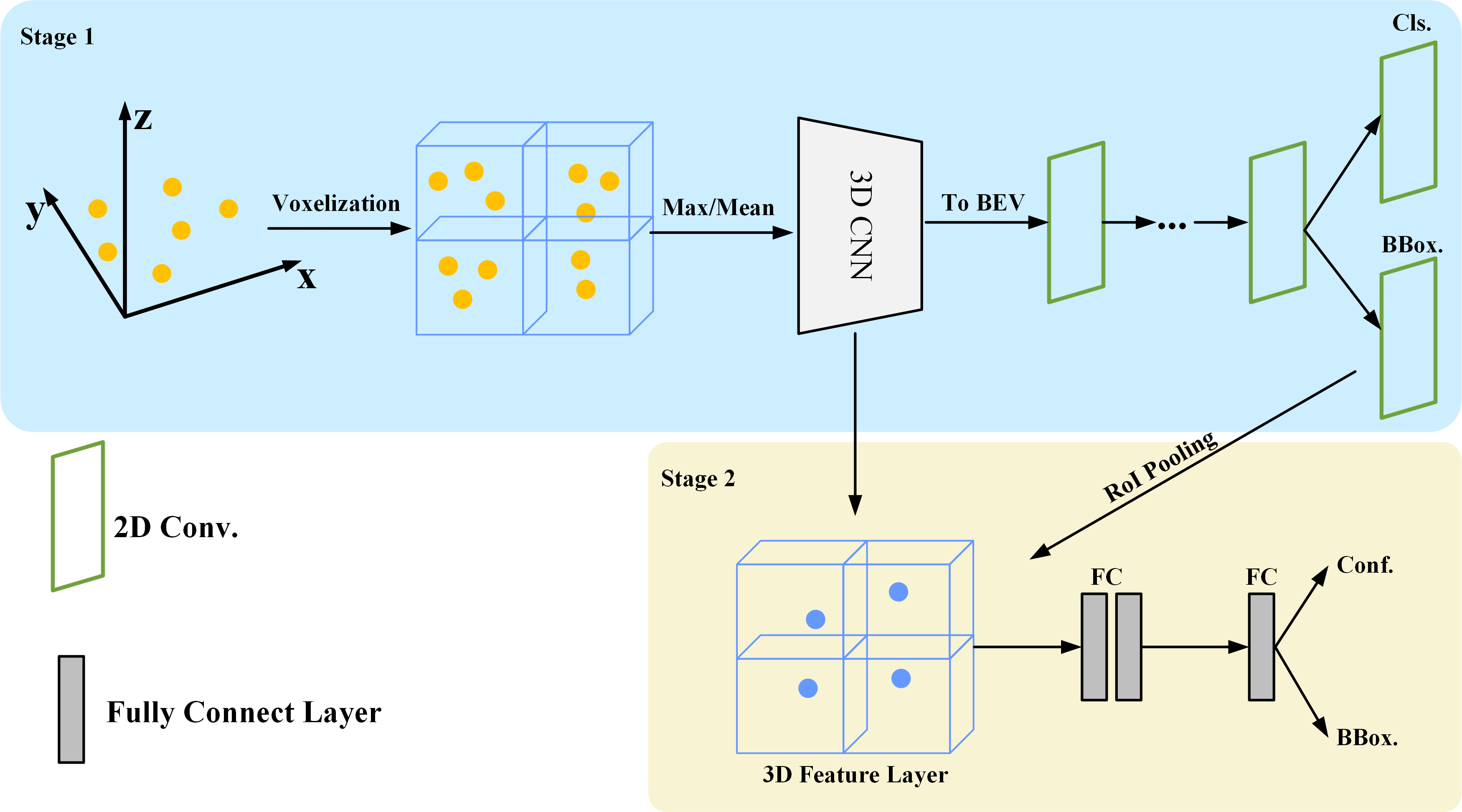}
	\caption{Overview of voxel-based 3D object detection methods. The blue box represents the basic structure of one-stage methods, while the combination of the blue and yellow box represents the basic structure of two-stage methods.}
	\label{fig_1}
\end{figure}

The voxel-based method is one of the mature methods for 3D object detection based on LiDAR. Its foundation lies in converting point clouds into structured voxel, followed by utilizing convolutional neural networks to accomplish the detection task \cite{zhou2018voxelnet, yan2018second}. As illustrated in \hyperref[fig_1]{Figure 1}, voxel-based methods can be classified into one-stage and two-stage methods. In the one-stage methods \cite{yan2018second,zhou2018voxelnet}, point clouds are voxelized, and 3D convolutional network is used for feature extraction. Then the extracted 3D features are squeezed into 2D bird's-eye view features, and 2D convolutional neural network is used for subsequent detection tasks. Although the one-stage methods have a simple structure and fast detection speed, it may result in the loss of vertical features when squeezing 3D features into 2D features. Otherwise, In the two-stage methods \cite{deng2021voxel}, region proposals, which are the bounding boxes generated in the first stage, are used to aggregate the 3D feature points generated by the 3D convolutional neural network. Then these aggregated features are used to accomplish the final detection task. Compared to the one-stage methods, the two-stage methods complete the detection task in 3D space, avoiding the features loss caused by squeezing 3D features into 2D features\cite{deng2021voxel}. This significantly enhances the detection accuracy of two-stage methods.

Although voxel-based two-stage methods have made significant progress, there are still some issues that need to be addressed: (1) In autonomous driving scenarios, for objects with fewer reflected points, such as cyclists, pedestrians, and distant objects, LiDAR can only obtain a limited amount of measurement data \cite{yin2021multimodal}. The point clouds generated by LiDAR are difficult to accurately describe the geometric features of these objects. In particular, voxelization and downsampling in 3D convolutional neural networks may lead to the loss of foreground feature points. For large objects like cars, the loss of a small number of feature points may not significantly affect the description of their geometric features in the 3D feature layer. However, for objects with fewer reflected points, even a small loss of feature points can greatly impact the description of their geometric features. This kind of geometric features aids the second stage in performing more precise bounding box regression. Therefore, the detection accuracy of voxel-based two-stage methods still performs poorly for objects with fewer reflective points. (2) The point clouds generated by LiDAR primarily reside on the surface of objects, so the centroid of a 3D object may be far from any surface point (the hollowness of point clouds) \cite{deng2021multi, qi2019deep}. Therefore, if a small radius is used for feature aggregation on the 3D feature layer, it may fail to aggregate all the foreground feature points. On the other hand, if a large radius is used for feature aggregation, it may aggregate the feature points that belong to other objects or the background, especially when the quality of region proposals is poor. However, previous voxel-based methods did not effectively address the issue of the hollowness of point clouds. To address the two aforementioned issues, we propose a novel two-stage voxel-based method called MS$^{2}$3D.

To improve the detection accuracy for objects with fewer reflective points, we construct a relatively compact 3D feature layer. We use a multi-branch voxel feature extraction method to extract voxel features at different scales. Then, these features from different branches are used to construct the 3D feature layer, incorporating features from different levels. Compared to the previous methods, our method results in a denser distribution of feature points in the 3D feature layer, encompassing various semantic feature features at multi-scales. Additionally, to avoid the loss of foreground feature points caused by downsampling, we propose a sampling method based on distance weights. We sample features based on the distance between feature points and the centroid of object, so as to preserve foreground points as much as possible. By combining these two methods, we have provided the 3D feature layer with abundant feature points, enabling it to better describe the geometric features of objects.

To address the hollowness of point clouds, we predict the offsets between the feature points in the 3D feature layer and the object's centroid, and then add these offsets to the coordinates of the feature points in three-dimensional space. This operation allows these feature points to be as close as possible to the object's centroid, enabling the aggregation of more foreground feature points even when using a smaller radius during feature aggregation. However, moving all feature points of an object to its centroid results in the loss of its original geometric features. Considering that feature points from deep-level have a larger receptive field and rich semantic features, we hope to aggregate all these points to participate in subsequent detection tasks. As for the feature points from shallow-level, they have smaller receptive fields and retain more geometric features. We would like to preserve them on the surface of objects to describe the geometric features of the objects. Therefore, we predict the offsets for the feature point clouds from deep-level, making them as close as possible to the object's centroid. On the other hand, for feature points from shallow-level, we retain them on the object's surface. Using this method, we can aggregate deep-level feature points rich in semantic features, while fully describing the geometric features of objects with shallow-level feature points. Subsequent experiments also demonstrate that our proposed feature aggregation method preserves the geometric features of objects in the 3D feature layer while improving the efficiency of 3D feature aggregation.

The primary contributions of this paper are outlined below:
\begin{itemize}
	\item{We propose a method for constructing a 3D feature layer using multi-branch voxel features and a method for sampling based on distance weights. So we can construct a denser 3D feature layer containing richer semantic features. Through the distance-weighted sampling method, we can preserve foreground feature points as much as possible during the feature extraction process. These two methods obviously enhance the detection accuracy of objects with fewer reflected points.}
	\item{We propose a method to address the impact of the hollowness of point clouds on feature aggregation. Through this method, we can aggregate as many deep-level feature points rich in semantic features as possible. On the other hand, we can preserve the geometric features of objects as much as possible.}
	\item{We conduct experiments on the KITTI and ONCE datasets to evaluate our method, demonstrating its robust performance even in complex environments with large-scale datasets.}
\end{itemize}
\section{Related Work}
3D object detection methods using LiDAR point clouds can be categorized based on their data representation formats: voxel-based, point-based, and projection-based methods. Furthermore, fusion methods that combine voxel-based and point-based methods also be proposed.
\subsection{Voxel-based Method}
Voxel-based methods typically transform point clouds into regular voxels to perform detection tasks. Currently, voxel-based methods can be categorized into one-stage and two-stage methods based on their network structures.

\textbf{One-Stage Methods:} 
VoxelNet \cite{zhou2018voxelnet} is the first one-stage voxel-based method. It first voxelizes the point clouds, transforming the point clouds into a series of voxels. Then, it performs feature extraction on the voxels using 3D convolution. Finally, the extracted 3D features are compressed into 2D bird's-eye view features, and the 3D detection task is completed from a bird's-eye view. The voxelization method introduced by VoxelNet enables the application of convolutional neural networks in 3D object detection based on LiDAR. However, in autonomous driving scenarios, the voxels obtained from point clouds are mostly zero-padded, resulting in a significant computational burden when performing feature extraction using 3D convolutional neural networks. To address this issue, SECOND \cite{yan2018second} applies sparse convolution for feature extraction, greatly improving the detection efficiency. In response to the misalignment issue between regression confidence and classification confidence, CIA-SSD \cite{zheng2021cia} proposes the IoU-aware confidence rectification module, while SA-SSD \cite{he2020structure} proposes an auxiliary network that can be turned off at any time to learn the 3D structural features of point clouds. In the end, CIA-SSD and SA-SSD both achieve a balance between regression confidence and classification confidence. The voxelization method proposed in VoxelNet introduces uncertainty in the embedded feature points within the voxels, which affects the final detection results. MVF \cite{zhou2020end} introduces a dynamic voxel encoding method to improve the stability of the detection results. In voxel-based methods, the design of voxel size has a significant impact on detection accuracy. Smaller voxels can provide more fine-grained features but impose a heavy computational burden and have limited receptive fields. Larger voxels can have a larger receptive field but cannot obtain precise position features. HVNet \cite{ye2020hvnet} and SMS-Net \cite{liu2022sms} adapt multiple scales of voxels as inputs, achieving a balance between fine-grained features and a large receptive field.

\textbf{Two-Stage Methods:} Two-stage methods further process 3D feature points based on the detection results of one-stage methods. Two-stage methods fundamentally achieve the detection task in 3D space rather than from the 2D bird's-eye view. PartA$^{2}$ \cite{shi2020points} is a two-stage method based on part-awareness, consisting of a part-aware stage and a part-aggregation stage. In the part-aware stage, PartA$^{2}$ generates 3D region proposals and accurate intra-object part locations. In the part-aggregation stage, RoI-aware Point Cloud Pooling is introduced to refine the region proposals by exploring the spatial relationships within the object parts. Utilizing the spatial features among various parts within the object, PartA$^{2}$ has achieved favorable detection results. However, the complex structure and additional part-awareness in PartA$^{2}$ impose computational burdens. Voxel RCNN \cite{deng2021voxel} proposes a relatively simple two-stage structure and proposes the Voxel RoI Pooling module, which directly aggregates 3D features on voxel features. This allows Voxel RCNN to achieve good accuracy even under coarse voxel granularity conditions. Additionally, the simple design of Voxel RCNN reduces computational costs. In general, two-stage methods directly perform the detection task in 3D space and can achieve higher regression accuracy. However, sparse 3D feature layer cannot effectively describe the geometric features of objects, and the hollowness of point clouds also pose challenges for feature aggregation.
\subsection{Point-based Method}
Unlike voxel-based methods, point-based methods typically utilize PointNet \cite{qi2017pointnet} and its variants \cite{thakur2020dynamic, yue2019dynamic,qi2017pointnet++,liu2019relation} for directly extracting features from point clouds. Point RCNN \cite{shi2019pointrcnn} uses PointNet++ \cite{qi2017pointnet++} for feature extraction and conducts the initial detection with the extracted features. Subsequently, the initial detection results are used to aggregate features for the points. Finally, the detection task is completed using the aggregated 3D features. VoteNet \cite{qi2019deep} proposes a deep Hough voting module for predicting instance centroid and has achieved promising results. 3DSSD \cite{yang20203dssd} reduces computational costs by removing all upsampling modules and proposes a fusion sampling method that enables detection on fewer representative points. IA-SSD \cite{zhang2022not} proposes two learnable, task-oriented instance-aware subsampling strategies that hierarchically select foreground points belonging to object-relevant regions of interest.
\begin{figure*}
	\centering
	\includegraphics[width=7.25in]{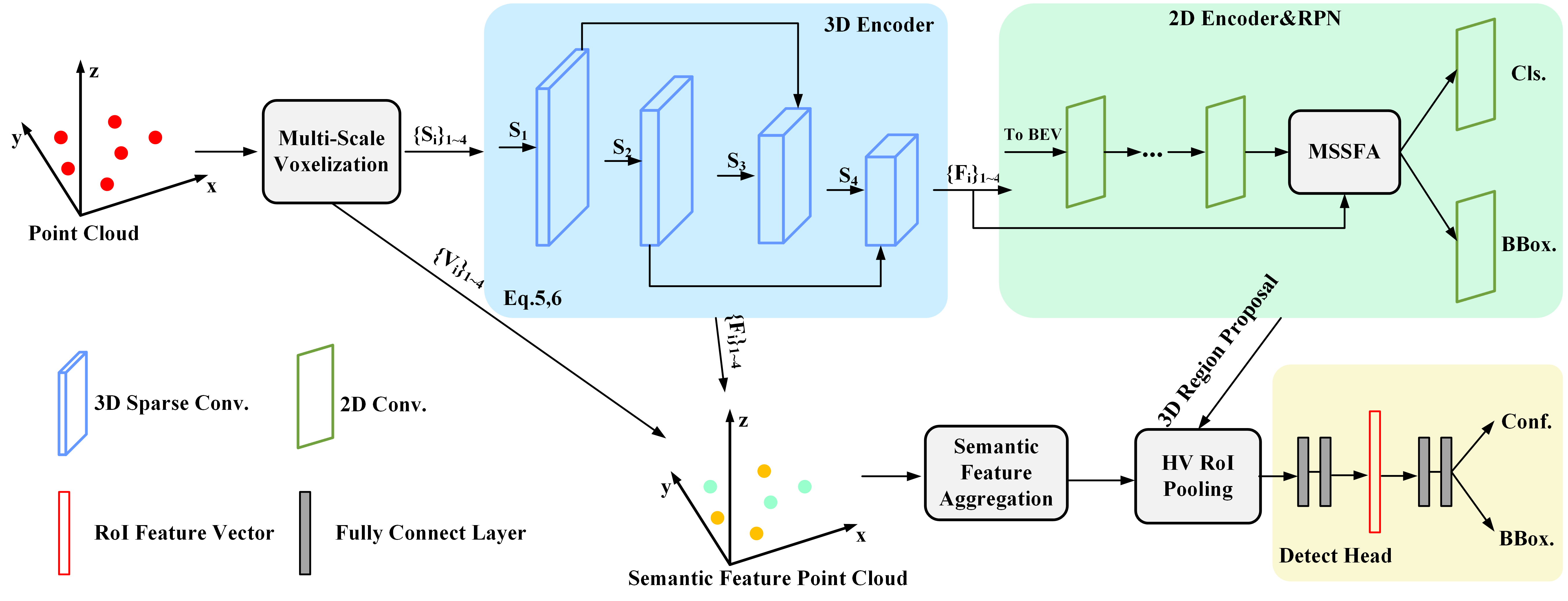}
	\caption{Overview of the MS$^{2}$3D structure. Here, $\left \{S_{i}\right \} _{1-4}$  denotes the Sparse Voxel Features of different sizes generated by the Multi-Scale Voxelization. Correspondingly,$\left \{F_{i}\right \}   _{1-4}$ represents the Sparse Voxel Features generated by the 3D encoder, and $\left \{V_{i}\right \} $$_{1-4}$ denotes the Voxel-wise Feature generated by the Multi-Scale Voxelization.}
	\label{fig_2}
\end{figure*}
\subsection{Projection-based Method}
Projection-based methods typically project point clouds into single or multiple views, converting them into structured 2D pseudo-image representations \cite{10454580, naqvi2023adversarial}. Subsequently, feature extraction is performed using 2D CNN. MV3D \cite{chen2017multi} projects point clouds into bird's-eye view and perspective view. It extracts features from the projection map and then fuses these features with the corresponding scene images. Complex YOLO and YOLO3D \cite{simony2018complex,ali2018yolo3d} conduct feature extraction and subsequent detection tasks from bird's-eye view. PointPillars \cite{lang2019pointpillars} encodes point clouds into top-down pillars from bird's-eye view with high-dimensional features. H$^{2}$3D RCNN \cite{deng2021multi} uses a feature encoding method akin to that of PointPillars. It effectively leverages features from both bird's-eye view and perspective view to acquire region proposals using 2D CNN.
\subsection{Voxel \& Point-based Method}
In order to surpass the limitations of voxel-based and point-based methods \cite{liu2019point}, some studies employ voxel-point representations for 3D detection tasks. PV-RCNN \cite{shi2020pv} merges the efficient feature extraction of 3D voxel CNN with the flexible receptive field of point-based networks, resulting in promising outcomes. HVPR \cite{noh2021hvpr} introduces an efficient memory module to enhance point-based features, achieving a commendable balance between accuracy and efficiency. BADet \cite{qian2022badet} develops a lightweight region aggregation network through local neighborhood graphs to attain more precise bounding box predictions.
\section{Methodology}
In this section, we will introduce the specific structure of MS$^{2}$3D and its corresponding loss function design. MS$^{2}$3D is a two-stage 3D object detection framework based on voxel. As illustrated in \hyperref[fig_2]{Figure 2}, MS$^{2}$3D comprises the following components: (1) Multi-Scale Voxelization, (2) 3D encoder, (3) 2D encoder, and (4) Semantic Feature Aggregation. The Multi-scale Voxelization divides the raw point clouds input into voxels of different scales. These voxels are subsequently input into the 3D encoder, producing sparse voxel features. Subsequently, the sparse voxel features are transformed into the bird's-eye view representation, and 3D region proposals are generated using the 2D encoder and the Region Proposal Network (RPN). In the 2D encoder, we utilize the MSSFA module proposed in our previous work PV-SSD \cite{10509820}, designed to facilitate the interaction between local and global features. We then integrate the feature points generated by Multi-scale Voxelization and the 3D encoder to construct the 3D feature layer. Finally, we employ the Hierarchical Voxel RoI Pooling (HV RoI Pooling) proposed in H$^{2}$3D R-CNN \cite{deng2021multi} to aggregate RoI features and refine the previously generated 3D region proposals. 
\subsection{Multi-Scale Voxelization}
Given the sparsity of the point clouds and the exacerbation of sparsity through voxelization, many voxel-based two-stage methods conduct aggregation within relatively sparse 3D feature layer. To tackle this issue, we propose a method to construct 3D feature layer using multi-branch feature points. We design the Multi-Scale Voxelization module to generate sparse voxel features with different voxel sizes, facilitating the subsequent construction of the 3D feature layer (notably, we do not simultaneously transform the input point clouds into voxels of different scales; instead, we first encode voxel features at smaller scales, and then perform large-scale voxelization using the features obtained from the smaller-scale encoding). Furthermore, in Multi-Scale Voxelization, we propose a method to sample feature points based on distance weights. This approach  helps mitigate the loss of foreground feature points during downsampling to some extent. To ensure the computed weight values are associated with the distance from the point clouds to the object's centroid, we design a loss function that serves as a supervisory signal for generating the distance weights.

During the process of Multi-Scale Voxelization, we voxelize the raw point clouds based on voxel sizes $\left \{s_{i}\right \} _{1-4}$. It's important to note that in the voxel encoding process, random sampling and zero-padding operations, as seen in hard voxelization \cite{zhou2018voxelnet}, may result in a significant loss of point clouds and uncertain embedding of point clouds. Therefore, in MS$^{2}$3D, we employ the dynamic voxelization \cite{zhou2020end}. The Feature encoder is utilized to obtain Sparse Voxel Feature $\left \{S_{i}\right \} _{1-4}$ and Voxel-wise Feature $\left \{V_{i}\right \} _{1-4}$. Consider that each additional size of feature input adds the computational consumption and reduces detection efficiency.  On the other hand, larger voxel sizes may lead to the inclusion of feature points from different objects within the same voxel, potentially affecting the final detection task. Therefore, we use four different sizes $i\in [1,4]$.

For instance, we employ voxelization with a size of $(0.1m, 0.1m,0.1m)$ for the KITTI dataset, resulting in subsequent voxel sizes of $(0.2m,0.2m,0.2m)$, $(0.4m,0.4m,0.4m)$, and $(0.8m,0.8m,0.8m)$ in the Multi-Scale Voxelization. The voxel sizes $(0.1m,0.1m,0.1m)$ and $(0.2m,0.2m,0.2m)$ correspond to small-scale voxels, containing local fine-grained features (the relative positional relationships between individual voxels contain the corresponding geometric features of the object). The voxel sizes $(0.4m,0.4m,0.4m)$ and $(0.8m,0.8m,0.8m)$ represent large-scale voxels, allowing for better capture of semantic features related to various components of an object. For example, the $(0.8m,0.8m,0.8m)$ voxels can capture semantic features about car tires and other components.
\begin{figure}[!t]
	\centering
	\includegraphics[width=3.5in]{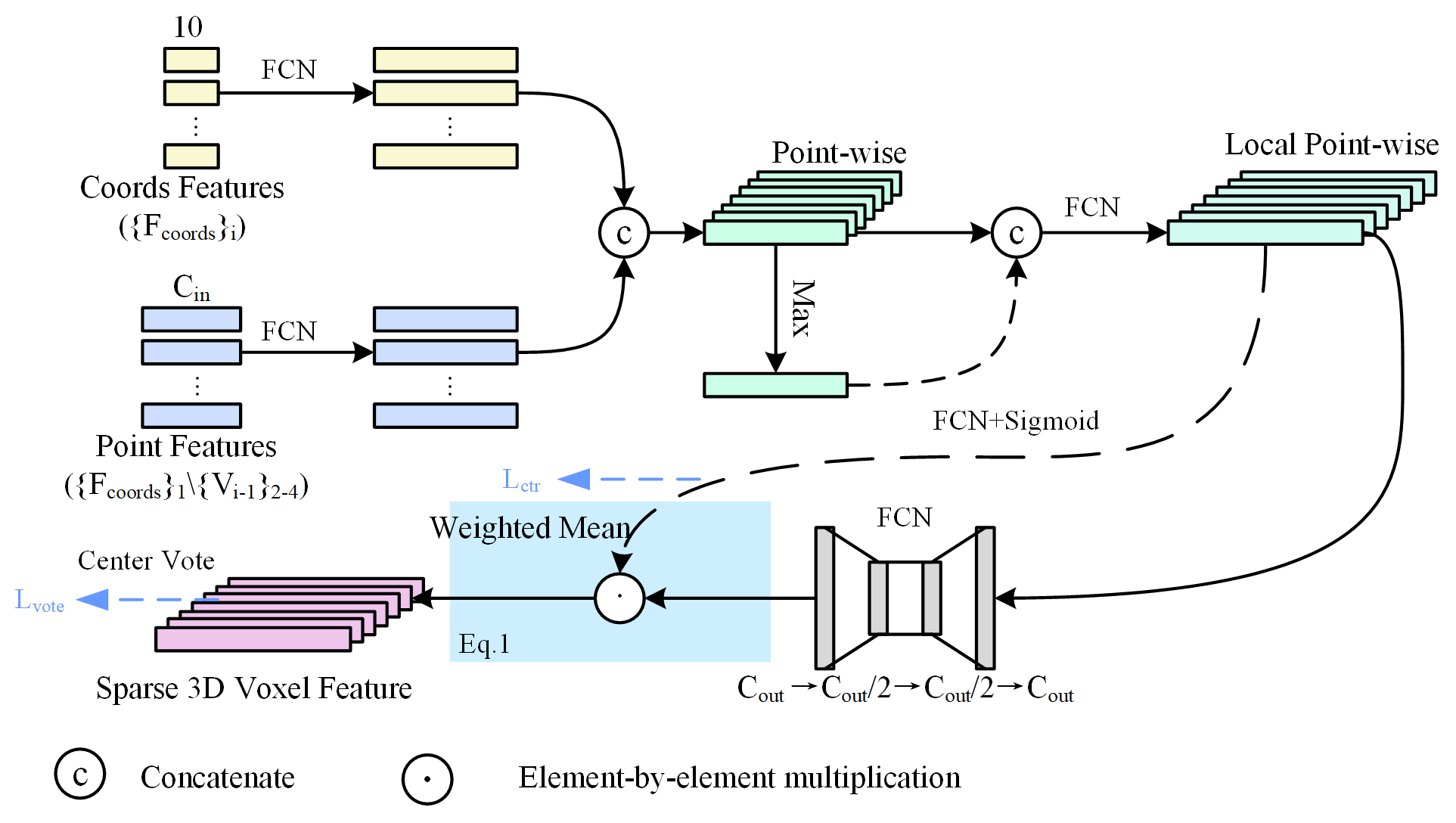}
	\caption{Overview of the structure of the feature encoding process in Multi-Scale Voxelization. The Center Vote operation and the calculation of $L_{vote}$ are only applied in the Multi-Scale Voxelization with the largest voxel size. The blue region in the figure represents the Weighted Mean module.}
	\label{fig_3}
\end{figure}

The detailed feature encoding method in Multi-Scale Voxelization is depicted in \hyperref[fig_3]{Figure 3}. The features of the point clouds, input to the feature encoder, are defined to comprise the coordinate features $C\in R^{N \times 10}$ and the point features $P\in R^{N \times C_{in}}$ (where $N$ is the number of points and $C_{in}$ is the number of input feature channels). The coordinate feature is represented as a 10-dimensional vector: $\left \{F_{coords}\right \}_{i}: (x, y, z, r, x_{c}, y_{c}, z_{c}, x_{p}, y_{p}, z_{p})$. Here, $x, y, z$ ,and $r$ denote the coordinates and reflectance intensity of the point clouds. Additionally, $x_{c}, y_{c}$, and $z_{c}$ represent the geometric center of all points in the voxel containing the point clouds; $x_{p}, y_{p}$ ,and $z_{p}$ are defined as $x-x_{c}, y-y_{c}, z-z_{c}$, respectively, indicating the point's relative position to the geometric center. For voxels of size $s_{1}$, point features and coordinate features are shared, meaning that the generated 10-dimensional vector $\left \{F_{coords}\right \}_{1}$ is used simultaneously for both point features and coordinate features for feature encoding, resulting in $V_{1}$. For subsequent voxels of size $\left \{s_{i}\right \} _{2-4}$, the $\left \{V_{i-1}\right \} _{2-4}$ is used for the point features.

To enhance the point cloud features, we use two fully connected network (FCN) layers to expand the dimensions of $C$ and $P$. This is followed by a concatenation operation in the feature dimension, resulting in the Point-wise Feature. Next, we apply a max operation to the Point-wise Feature, followed by concatenation, resulting in the Local Point-wise Feature $P_{local}\in R^{N \times C^{'}}$. We use an FCN followed by a Sigmoid function to generate distance weights $W_{d}\in R^{N \times 1}$, describing the distance of each point from the object's centroid ($W_{d}\in[0,1]$, where values closer to 1 indicate proximity to the object's centroid). The $P_{local}$ is integrated using two FCN layers. In the Weighted Mean module, we perform element-wise multiplication between $W_{d}$ and the integrated $P_{local}$, resulting in Voxel-wise Feature $V_{i}\in R^{N \times C_{out}}$. Next, for $V_{i}$ with the same voxel index, we perform the mean operation to obtain $S_{i}$. We select the top K $V_{i}$ with the highest weights based on $W_{d}$. These selected features become the input for the next Multi-Scale Voxelization module. In autonomous driving scenarios, the point clouds  often contains a significant number of background points. For the task of 3D object detection, prioritizing point cloud features closer to the objects' centroid is crucial. The Weighted Mean operation gives more attention to the point cloud features closer to the object's centroid. The coordinates of $V_{i}$ (indexed from 1 to 4) are defined as $X_{i}\in R^{N \times 3}$. The equation for the Weighted Mean operation is shown as \hyperref[eq1]{Eq. (1)}.
\begin{equation}
	\label{eq1}
	mean_{w} (x,W_{d}, X_{i})=mean(W_{d} \cdot x, X_{i})
\end{equation}

Furthermore, inspired by VoteNet \cite{qi2019deep}, we hope to make the feature points with rich semantic features closer to the object's centroid. Therefore, in the final layer of the Multi-Scale Voxelization feature encoder, we introduce the Center Vote operation, which predicts the offset of Voxel-wise Feature from the object's centroid. Specifically, we use a fully connected layer to change the number of channels of the feature points to 3, corresponding to the offset components of the feature points in the x, y, and z directions relative to the object's centroid. Then, we directly add these three offset components to the coordinates of the corresponding feature points, moving them closer to the object's centroid. Additionally, to avoid generating large offsets in a single step that would shift the feature points far from the object's centroid, we limit the maximum offsets in the x, y, and z-axis to $(3m, 3m, 2m)$, respectively. We use $L_{vote}$ and $L_{ctr}$ to supervise the prediction of offsets from the Center Vote, as well as the prediction of $W_{d}$. The specific equations are presented as \hyperref[eq2]{Eq.(2)}, \hyperref[eq3]{(3)}, and \hyperref[eq4]{(4)}.

\begin{equation}
	\label{eq2}
	Mask_{i}=\sqrt[3]{\frac{min(f^{*}, b^{*})}{max(f^{*}, b^{*})}\times \frac{min(l^{*}, r^{*})}{max(l^{*}, r^{*})}\times \frac{min(u^{*}, d^{*})}{max(u^{*}, d^{*})}}
\end{equation}

\begin{equation}
	\label{eq3}
	L_{ctr}=-(Mask_{i}\cdot a_{i}log(W_{d})+(1-a_{i})log(1-W_{d}))
\end{equation}
\begin{equation}
	\label{eq4}
	L_{vote}^{v}=\frac{1}{M_{pos}}\sum \mathbb 1_{pos}\left \|\bigtriangleup x_{i}-\bigtriangleup \hat{x}_{i}\right\|
\end{equation}
where $f^{*},b^{*},l^{*},r^{*},u^{*},d^{*}$ respectively denote the distances of feature points to the six surfaces (front, back, left, right, up, and down) of the ground truth bounding box. $a_{i}$ denotes the one-hot encoding of foreground and background points in the point clouds. $M_{pos}$ denotes the count of foreground points. $\bigtriangleup x_{j}\in R^{3}$ denotes the coordinates of the object's centroid, while $\bigtriangleup \hat{x}_{j} \in R^{3}$ denotes the new coordinates of the feature points obtained by adding the predicted coordinate offsets. $Mask_{i}$ are defined in \hyperref[eq2]{Eq.(2)}. During the training phase, using $Mask_{i}$ in \hyperref[eq2]{Eq.(3)} can assign different weights to feature points based on their distances from the object centroid (with points outside the ground truth bounding box assigned a weight of 0), implicitly incorporating geometric priors into the network training \cite{zhang2022not}. By incorporating $M_{pos}$ into the \hyperref[eq4]{Eq.(4)}, interference from background points on weight prediction can be reduced. As for \hyperref[eq4]{Eq.(4)}, the value of $\mathbb 1_{pos}$ is 1 for foreground feature points and 0 for background feature points. This ensures that only foreground feature points are considered during the computation of $L_{vote}^{v}$, preventing irrelevant background feature points from being shifted close to positive samples or even onto the object's interior.
\subsection{3D Encoder}
In the 3D encoder, we utilize sparse convolution, as proposed in SECOND \cite{yan2018second}, for further feature extraction on  $\left \{S_{i}\right \} _{1-4}$. We define $\left \{f_{i}\right \} _{1-4}$ as the sparse convolution corresponding to $\left \{S_{i}\right \} _{1-4}$. To optimize computational efficiency in the subsequent 2D encoder, we apply different levels of downsampling to $S_{i}$. Specifically, $S_{1}$ and $S_{2}$ are downsampled by a factor of 4, and for $S_{3}$, we perform a 4$\times$ downsampling in the $W$ and $H$ dimensions and a 2$\times$ downsampling in the $D$ dimension. For $S_{4}$, we perform a 4$\times$ downsampling in the $W$ and $H$ dimensions, but no downsampling in the $D$ dimension (note that during the computation of sparse convolution, $S_{i}$ will be reconstructed as $S_{i} \in R^{W_{i} \times H_{i} \times D_{i} \times C_{i}}$). Each $f_{i}(x)$ is composed of 2 blocks, and each block consists of a  $3 \times 3$ submanifold convolution\cite{yan2018second} and two 3×3 sparse convolutions\cite{yan2018second}. In each block of $f_{1}(x)$ and $f_{2}(x)$, the submanifold convolution performs a $2 \times$ downsampling, i.e., $stride=2$, $padding=1$. In $f_{3}(x)$ and $f_{4}(x)$, only the submanifold convolution in the first block performs $2 \times$ downsampling, i.e., $stride=2$, $padding=1$. The remaining convolutions do not perform downsampling, i.e., $stride=1$, $padding=1$. Additionally, the submanifold convolution in the first block of $f_{i}(x)$ will expand the number of channels of the input features by a factor of 2, i.e., the output channel number of the submanifold convolution is $ 2 \times C_{i}$. Furthermore, to enhance the richness of the acquired semantic features, we pass the outputs of $f_{1}(x)$ and $f_{2}(x)$ to $f_{3}(x)$ and $f_{4}(x)$ respectively, and concatenate them with the outputs of $f_{3}(x)$ and $f_{4}(x)$. The process of feature encoding in the 3D encoder is as follows.
\begin{equation}
	\label{eq5}
	\left \{ F_{i} \right \}_{1,2}=f_{1,2}(S_{1,2})
\end{equation}
\begin{equation}
	\label{eq6}
	\left \{ F_{i} \right \} _{3,4}=Cat(F_{1,2},f_{3,4}(S_{3,4}))
\end{equation}
\subsection{2D Encoder}
\begin{figure}[!t]
	\centering
	\includegraphics[width=3.5in]{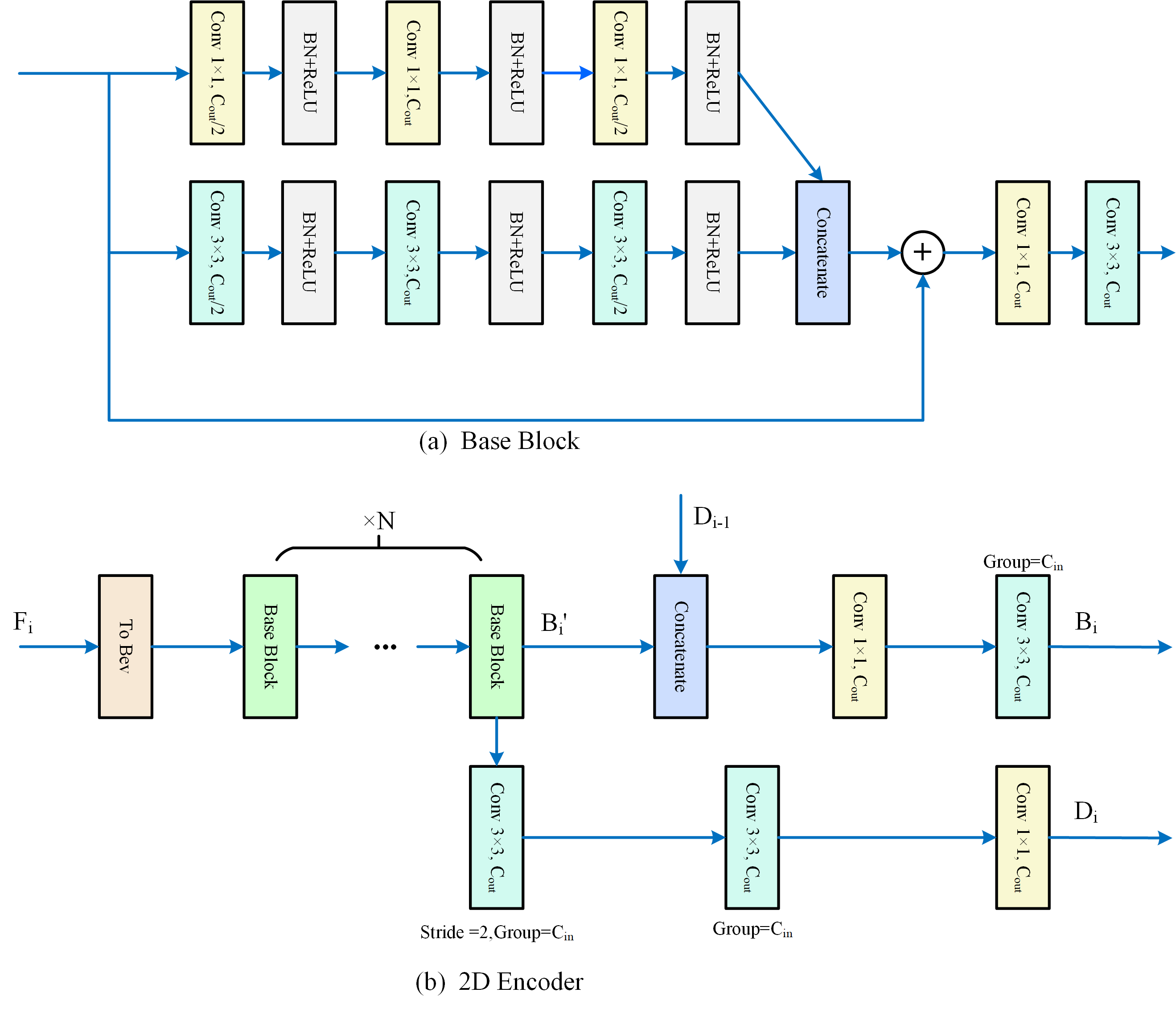}
	\caption{Overview of the 2D encoder structure. In (a), we depict the architecture of the Base Block within the 2D encoder. Here, BN denotes the Batch Normalization operation. In (b), we present the overall structure of the 2D encoder. Notably, for $i=4$, there is no downsampling branch.}
	\label{fig_4}
\end{figure}
In the 2D encoder, we start by transforming $\left \{ F_{i} \right \} _{1-4}$ into feature maps in the bird's-eye view. We then perform further feature extraction using 2D convolutions to generate region proposals from the bird's-eye view in the subsequent steps. 

As shown in \hyperref[fig_4]{Figure 4 (b)}, we define the 2D feature obtained using $N$ Base Blocks as $B_{i}^{'} \in R^{W^{'} \times H^{'} \times C^{'}}$. Next, $B_{i}^{'}$ undergoes two $3 \times 3$ convolution layers and a $1 \times 1$ convolution layer for 2$\times$ downsampling, resulting in a downsampled 2D feature $D_{i} \in R^{\frac{W^{'}}{2}\times \frac{H^{'}}{2} \times C^{'}}$ (we use $3 \times 3$ convolution layer and $1 \times 1$ convolution layer to transform the 3D features into 2D features in the bird's-eye view; for downsampling, we employ a $3 \times 3$ convolution with $stride=2$, $padding=1$ and $group= C_{in}$; it's important to note that no downsampling is performed when $i=4$). Additionally, we concatenate the output $D_{i-1}$ from the previous layer with $B_{i}^{'}$ to ensure that the output feature contains richer semantic features. Finally, feature integration is performed using a $1 \times 1$ convolution layer and a $3 \times 3$ convolution layer, yielding the 2D bird's-eye view features $\left \{ B_{i}  \right \}_{1-4} \in R^{W^{'} \times H^{'} \times C^{'}}$.

In the Base Block of the 2D encoder, we design a feature encoding structure considering both spatial and feature dimensions. As shown in \hyperref[fig_4]{Figure 4 (a)}, the Base Block comprises the $1 \times 1$ branch, the $3 \times 3$ branch, and the residual branch. The $1 \times 1$ branch integrates features across the feature dimension of the 2D feature map, while the $3 \times 3$ branch integrates features across the spatial dimension to capture local features between neighboring pixels. The outputs of the $1 \times 1$ and $3 \times 3$ branches are concatenated and element-wise added to the output of the residual branch. Subsequently, further feature integration is performed using a $1 \times 1$ convolution layer and a $3 \times 3$ convolution layer. Both the $1 \times 1$ and $3 \times 3$ branches consist of three sets of Convolution + ReLU + Batch Normalization combinations.
\subsection{Semantic Feature Aggregation}
Currently, LiDAR typically captures only a few measurements for small or distant objects. Consequently, accurately describing the geometric features of these objects, especially in the 3D feature layer, can be challenging. To tackle this issue, we propose Semantic Feature Aggregation to construct a relatively compact 3D feature layer using semantic feature points generated during the feature encoding process at different levels.
\begin{figure}[!t]
	\centering
	\includegraphics[width=3.5in]{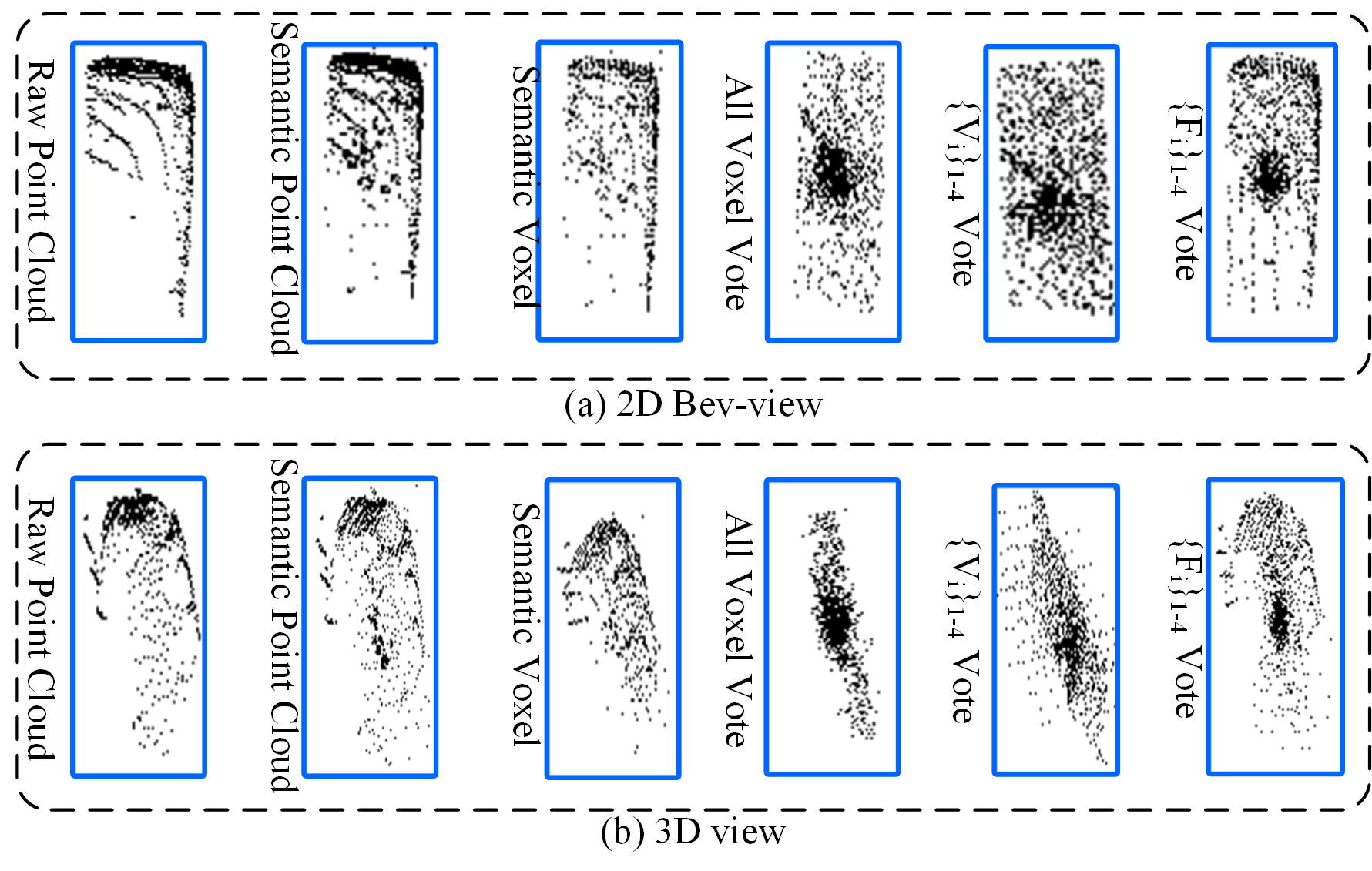}
	\caption{Overview of the semantic point clouds and the three Center Vote schemes. In (a), the results are visualized from the bird's-eye view, while in (b), the results are visualized in 3D perspective.}
	\label{fig_5}
\end{figure}

As shown in \hyperref[fig_5]{Figure 5} (‘Semantic Point Cloud’ represents the set of feature points constructed using different scales of semantic feature points. ‘Semantic Voxel’ represents the segmented point clouds after voxelization), the semantic point clouds, aggregated from different levels of semantic features, is denser than the raw point clouds. This enables a more accurate depiction of object's geometric features. Even after voxelization, the resulting sparse semantic voxel preserves more details than the raw point clouds. It's important to note that the implicit small-scale feature point offsets during the voxel coordinate transformation process and the center vote operation in the final layer of feature points within the Multi-Scale Voxelization module prevent a significant number of feature points from having identical coordinates.

Additionally, due to the hollowness of the point clouds, effectively aggregating features in the second phase of feature aggregation within the 3D feature layer becomes challenging. Specifically, when the region proposals generated in the first stage lack precision, feature aggregation with a large radius may merge irrelevant background points. On the other hand, utilizing a small radius for feature aggregation might fail to encompass all foreground points. This difficulty in feature aggregation makes it challenging to accurately regress the bounding box from the RPN during box refinement.

Inspired by VoteNet \cite{qi2019deep}, we apply the Center Vote operation to the voxelized semantic voxel to predict the offset from the object's centroid (the Center Vote operation is the same as the one used in the Multi-Scale Voxelization module). This allows the semantic voxel to closely approximate the object's centroid. We present three schemes for Center Vote (scheme 3 is the one employed in MS$^{2}$3D):
\begin{itemize}
	\item{Center Vote for all semantic voxels}
	\item{Center Vote only for $\left \{ V_{i} \right \}_{1-4}$ encoded through Multi-Scale Voxelization.}
	\item{Center Vote only for $\left \{ F_{i} \right \}_{1-4}$ encoded through the 3D encoder.}
\end{itemize}

The visualization of the 3D feature layer constructed using these three schemes are shown in \hyperref[fig_5]{Figure 5}. In \hyperref[fig_5]{Figure 5 (a)} (‘All Voxel Vote’ represents scheme 1; ‘$\left \{ V_{i} \right \}{1-4}$ Vote' represents scheme 2; ‘$\left \{ F{i} \right \}_{1-4}$ Vote' represents scheme 3), when performing Center Vote for all semantic voxels, we observe that most feature points in the 3D feature layer are concentrated around the object's centroid. However, as depicted in \hyperref[fig_5]{Figure 5 (b)}, the 3D feature layer compresses in the vertical direction near the plane of the object's centroid when viewed from the 3D perspective. While this compression aids subsequent HV RoI Pooling in feature aggregation, it also causes the 3D feature points to lose the object's original geometric feature.

When performing Center Vote only for $\left \{ V_{i} \right \}_{1-4}$ in the 3D feature layer, we observe that although the feature points are relatively dispersed in the vertical direction, they still cannot accurately capture the object's geometry features.

By performing Center Vote solely for $\left \{ F_{i} \right \}_{1-4}$, the 3D feature layer aggregates feature points near the object's centroid while effectively representing the object's geometry. Furthermore, $\left \{ F_{i} \right \}_{1-4}$ are sparse voxel features encoded by a deep network, containing rich semantic features, while $\left \{ V_{i} \right  \}_{1-4}$ are sparse voxel features encoded by a shallow network, preserving the object's geometric feature. Maintaining $\left \{ V_{i} \right  \}_{1-4}$ on the object's surface enhances its geometric characterization. This construction of the 3D feature layer is particularly advantageous for subsequent box refinement. In the ablation experiments, we conduct a detailed analysis of these three schemes. It is important to note that for scheme (2) and scheme (3), after voxelization, some voxels may simultaneously contain $\left \{ V_{i} \right \}_{1-4}$ and $\left \{ F_{i} \right \}_{1-4}$. However, we perform the Center vote operation only on voxels that exclusively contain either $\left \{ V_{i} \right \}_{1-4}$ or $\left \{ F_{i} \right \}_{1-4}$.

\begin{figure}
	\centering
	\includegraphics[width=3.5in]{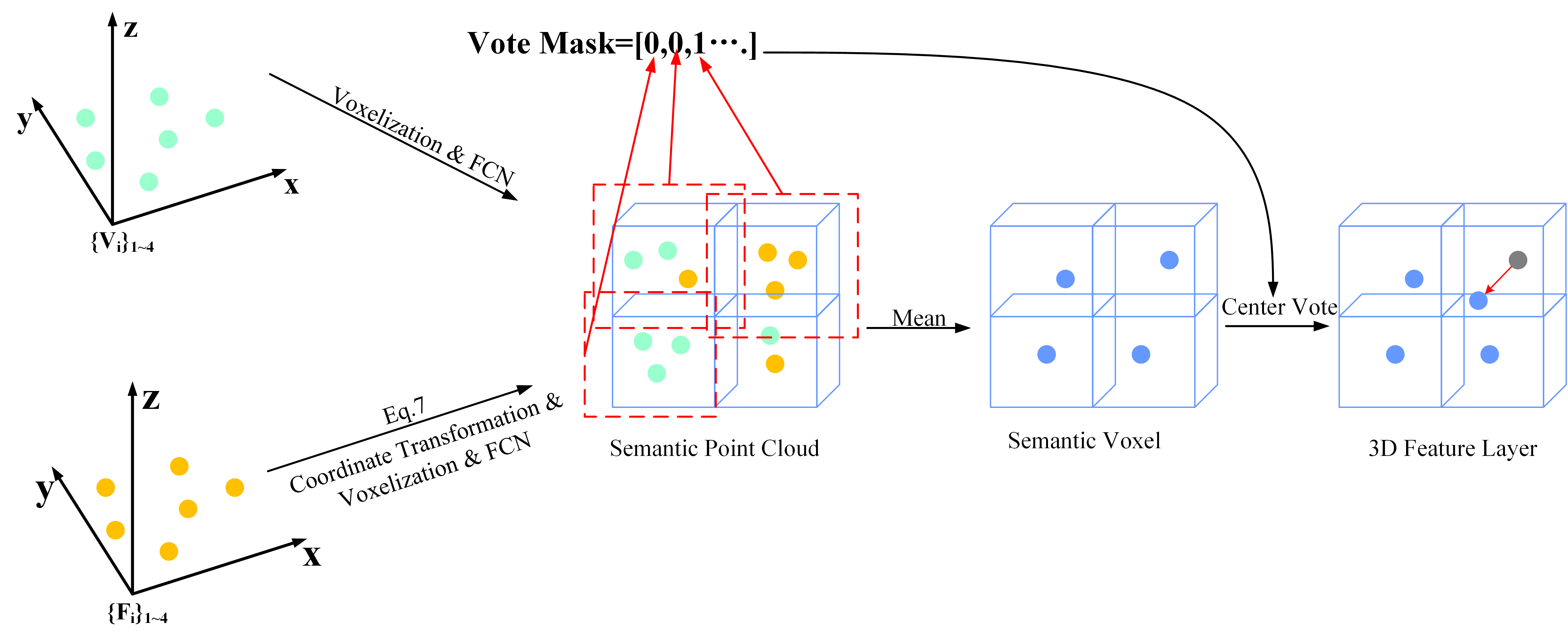}
	\caption{Overview of the Semantic Feature Aggregation Structure. The red arrows indicate the offsets generated by the Center Vote. The gray points denote the original positions, and the blue points indicate the positions after the offsets are applied.}
	\label{fig_6}
\end{figure}
\hyperref[fig_6]{Figure 6} illustrates the structure of Semantic Feature Aggregation. We define the coordinates of $\left \{ F_{i} \right \}_{1-4}$ as $X_{F}^{'} \in R^{N \times 3}$, where $X_{F}^{'}$ are coordinates in the voxel coordinate system. On the other hand, the coordinates of $\left \{ V_{i} \right \}_{1-4}$ are define as $X_{V}$ in the world coordinate system. First, we transform $X_{F}^{'}$ back to the world coordinate system using the coordinates of each voxel's center point, resulting in $X_{F} \in R^{N \times 3}$. The coordinate transformation formula is given by \hyperref[eq7]{Eq. (7)}. Subsequently, we adjust $\left \{ F_{i} \right \}_{1-4}$ and $\left \{ V_{i} \right \}_{1-4}$ to have the same feature dimension through two FCN layers, followed by concatenation to generate the Semantic Point Cloud. Then, for feature points in the generated Semantic Point Cloud with the same voxel index, we perform a mean operation to obtain the Semantic Voxel. Additionally, after generating the Semantic Point Cloud, we create a mask $Mask_{vote}$ based on whether a voxel contains $\left \{ V_{i} \right \}_{1-4}$ (0 if it contains $\left \{ V_{i} \right \}_{1-4}$, 1 otherwise). This mask is used for subsequent Center Vote. After applying Center Vote using the mask to the Semantic Voxel, we obtain the final 3D feature layer. HV RoI Pooling is then utilized for 3D feature aggregation, followed by box refinement using FCN. The formulation for Semantic Feature Aggregation is provided in \hyperref[eq8]{Eq. (8)}.

\begin{equation}
	\label{eq7}
	X_{F}=(X_{F}^{'}+\left [ 0.5,0.5,0.5 \right ] \times V_{size} + min(Range))
\end{equation}
\begin{multline}
	\label{eq8}
	F_{3D}=Vote(mean(Cat(FCN(\left \{ V_{i} \right \}_{1-4}), \\ FCN(\left \{ F_{i} \right \}_{1-4})),X) \times Mask_{vote})
\end{multline}
where $V_{size} \in R^{1\times 3}$ denotes the size of the voxel; $min(Range)$ denotes the minimum value of the range of the point clouds fed into the network; $X=\left \{ X_{F}, X_{V} \right \}$ denotes the coordinates of the Segment Point Cloud; $FCN(x)$ denotes the Fully Connected Layer; $Cat(x)$ denotes the Concatenate operation; $mean(x)$ denotes the mean operation; $Vote(x)$ is the Center Vote operation; $F_{3D}$ denotes the 3D feature layer.

Additionally, we have designed a loss function within the Semantic Feature Aggregation to supervise the offsets generated by the Center Vote. Similar to the method described in Multi-Scale Voxelization, this supervision is exclusively applied to foreground points, with background points excluded from participating in the loss calculation. However, in this context, we calculate the loss only for foreground points where $Mask_{vote}$ equals 1. The exact calculation formula is depicted in \hyperref[eq9]{Eq. (9)}.
\begin{equation}
	\label{eq9}
	L_{vote}^{f}=\frac{1}{M_{pos}} \sum Mask_{vote}\cdot a_{i}\left \| \bigtriangleup x_{i}-\bigtriangleup \hat{x}_{i}  \right \| 
\end{equation}

\subsection{Loss Function}
In the training phase, MS$^{2}$3D uses end-to-end optimization, and the total loss function is calculated as shown in \hyperref[eq10]{Eq. (10)}:
\begin{equation}
	\label{eq10}
	L_{total}=L_{rpn}+L_{head}+\alpha L_{vote}+\beta L_{ctr} 
\end{equation}
where $L_{rpn}$ is used to train the RPN \cite{yan2018second, zhou2018voxelnet}, $L_{head}$ is used to train the detection head \cite{deng2021voxel}, and $L_{vote}$ is used to train the Center Vote portion of Multi-Scale Voxelization feature encoding and Semantic Feature Aggregation. $\alpha = 1, \beta=0.25$. The specific calculation equation is as follows:
\begin{equation}
	\label{eq11}
	L_{rpn}=\frac{1}{N_{fg}}\left [ L_{focal}(x_{i},x_{i}^{*})+ \mathbb 1(x_{i}^{*}>1)\sum L_{loc}(y_{i},y_{i}^{*})\right ] 
\end{equation}
$N_{fg}$ denotes foreground anchors; $x_{i}$ and $y_{i}$ denote the outputs of the categorization branch and the bounding box regression branch; $x_{i}^{*}$ and $y_{i}^{*}$ denote the labels of the corresponding categorization and bounding box regressions; $\mathbb 1(x_{i}^{*}>1)$ denotes that only foreground anchors are involved in the computation of regression loss. $L_{loc}$ is Smooth L1 Loss, and $L_{focal}$ is Focal Loss.

In the detection head, the target confidence of the $ith$ region proposal is calculated based on the Intersection over Union (IoU) of its corresponding ground-truth box:
\begin{equation}
	\label{eq12}
	r_{i}^{*}(IoU_{i})=\left\{\begin{matrix}
		0 & IoU_{i}<\theta _{l}\\
		\frac{IoU_{i}-\theta _{l}}{\theta _{h}-\theta _{l}}  & \theta _{l} <IoU_{i}<\theta _{h}\\ 
		1&IoU_{i}>\theta _{h}
	\end{matrix}\right.
\end{equation}
where $\theta _{h}$ and $\theta _{l}$ denote the IoU thresholds for foreground and background, respectively. The loss in the detection head section is calculated as follows:
\begin{multline}
	\label{eq13}
	L_{head}=\frac{1}{N_{s}} [\sum L_{bce}(\gamma _{i},\gamma _{i}^{*}(IoU_{i}))+ \\  \mathbb 1(IoU_{i}\ge \theta _{reg}) \sum L_{reg}(z_{i},z_{i}^{*})] 
\end{multline}
$L_{bce}$ denotes Binary Cross Entropy Loss; $L_{reg}$ denotes Smooth L1 Loss; $\gamma_{i}$ denotes the predicted value of confidence; $z_{i}$ denotes the regression prediction of the detection head; $N_{s}$ denotes the number of region proposals sampled. $L_{vote}$ is computed as follows:
\begin{equation}
	\label{eq14}
	L_{vote}=L_{vote}^{v}+L_{vote}^{f}
\end{equation}
where the specific formulas for $L_{vote}^{v}$ and $L_{vote}^{f}$ are introduced in \hyperref[eq2]{Eq. (4)} and \hyperref[eq9]{Eq. (9)}.
\section{Experiments}
\subsection{Datasets and Evaluation Metrics}
\textbf{KITTI Dataset:} The KITTI dataset \cite{geiger2012we} is an open-source dataset designed for autonomous driving scenarios, containing road data collected from various sensors such as LiDAR and cameras. The dataset consists of 7,481 training samples and 7,518 testing samples. Typically, the training data is divided into a training set with 3,712 samples and a validation set with 3,769 samples. In this paper, we randomly split the training set into a training subset of 6,000 samples and a validation subset of 1,481 samples. The training subset is used for model training, while the validation subset assists in selecting the best-performing model. The chosen model is then evaluated on the KITTI testing set. Additionally, we use the data split of 3,712 training samples and 3,769 validation samples for ablation experiments. The KITTI dataset categorizes data into three difficulty levels (Easy, Mod., Hard) based on object size, occlusion, and truncation factors.

\textbf{ONCE Dataset:} To further assess the effectiveness of our method, we evaluate our method using the ONCE dataset \cite{mao2021one}. This dataset includes 6 sequences for training and 4 sequences for validation. We train our method on the training sequences and evaluate its performance on the validation sequences. The ONCE dataset classifies objects into four evaluation levels (overall, objects in the $0-30m$ range, objects in the $30-50m$ range, and objects $>50m$ away) based on their distance from the sensor. Evaluation is carried out using Average Precision (AP) and Mean Average Precision (mAP) as the metrics.
\subsection{Implementation Details}
\textbf{Voxelization:} For the KITTI dataset, we select point clouds within the cartesian coordinate range of $x=[0,70.4]m, y=[-40,40]m$ and $z=[-3.0,1.0]m$. The voxel size is set at $(0.1m,0.1m,0.1m)$. Following the approach in \cite{zhou2020end}, we employ dynamic voxelization, which avoids random point discarding or zero-padding within voxels. For the ONCE dataset, we consider point clouds within the cartesian coordinate range of $x=[-73.6,73.6]m, y=[-73.6,73.6]m$ and $z=[-5.0,3.0]m$. The voxel size is set at $(0.2m,0.2m,0.2m)$.
\begin{figure*}
	\centering
	\includegraphics[width=7.25in]{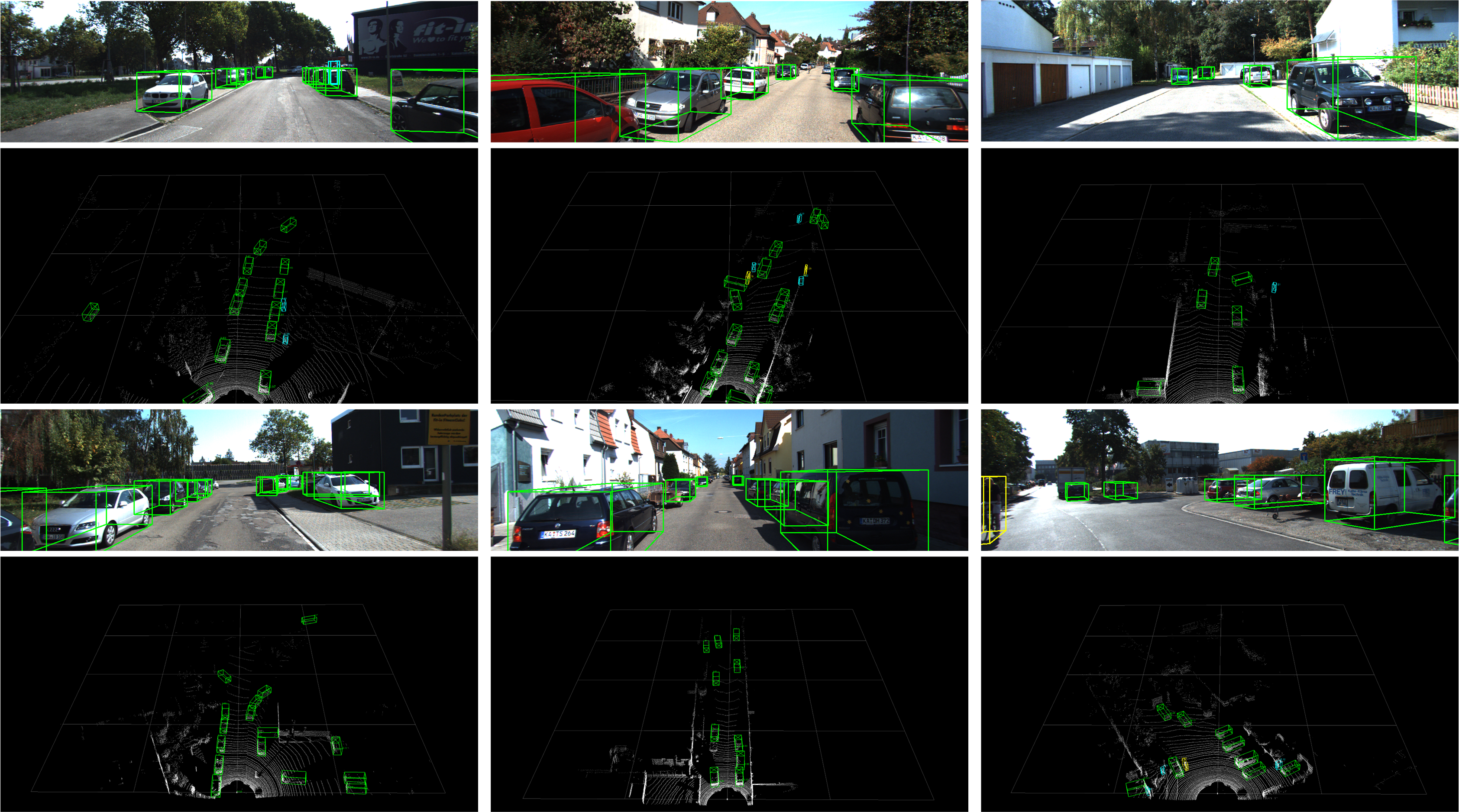}
	\caption{Visualization of detection results on the KITTI dataset. The first and third rows display the detection results projected onto images, while the second and fourth rows present the results on the point clouds. The green bounding boxes correspond to ‘Car’, the yellow bounding boxes represent ‘Cyclist’, and the blue bounding boxes indicate ‘Pedestrian’.}
	\label{fig_7}
\end{figure*}

\textbf{Data Augmentation:} We follow the methodology presented in SECOND \cite{yan2018second} by randomly selecting objects from a constructed sample library and adding them to the current sample. For the KITTI dataset, we randomly select 15 cars, 15 pedestrians, and 15 cyclists. For the ONCE dataset, we randomly select 1 car, 1 bus, 3 trucks, 2 pedestrians, and 2 cyclists. Additionally, object instances underwent random rotation (with rotation angles selected from $[- \pi /4, \pi/4]$), random scaling (scaling factors drawn from $[0.95, 1.05]$) ,and random horizontal flipping along the x-axis \cite{yan2018second,zhou2018voxelnet,yang2018pixor}.

\textbf{Network Architecture:} The feature encoder feature dimensions change as follows: first voxelization encoding $(10\longrightarrow 16\longrightarrow 16)$; second $(16\longrightarrow 24\longrightarrow 24)$; third $(24\longrightarrow 24\longrightarrow 32)$; fourth $(32\longrightarrow 32\longrightarrow 48)$. In Multi-Scale Voxelization module, the sampling rates is $[40\%,40\%,60\%]$ (where a sampling rate of 40\% means that the top 40\% of feature points with the largest $W_{d}$ values are selected). In the 2D encoder, each Base Block is repeated three times. The channel progression for the first two layers is $(96\longrightarrow 128\longrightarrow 256)$, and for the last two layers, it is $(128\longrightarrow 128\longrightarrow 256)$. HV RoI Pooling follows the same design as H$^{2}$3D RCNN \cite{deng2021multi}. Region proposals are divided into coarse grids $(3 \times 3 \times 3)$ and fine grids $(6 \times 6 \times 6)$. The query radius is 6 for coarse grids and 3 for fine grids. The MS$^{2}$3D model encoding parameters are shown in \hyperref[table1]{Table 1}. The parameters of FCN are the input and the number of output channels; FCN$_{P}$ and FCN$_{C}$ represent the FCN modules encoding point features and coordinate features in \hyperref[fig_3]{Figure 3}, respectively. The parameters for sparse convolution and submanifold convolution include the kernel size, padding, stride, and the number of output channels. The parameters for the Base Block are the number of output channels of the convolutions $C_{out}$ in \hyperref[fig_4]{Figure 4}; the Base Block structure is detailed in \hyperref[fig_4]{Figure 4}. $C^{v}_{i,0} \in [10, 16, 24, 32]$, $C^{v}_{i,1} \in [16,24, 24, 32]$, $C^{v}_{i,2} \in [16, 24, 32, 48]$. When $i=0$ or $1$, $stride=2$; for the others, $stride=1$.

\textbf{Training:} For training hyperparameters, we used the same hyperparameter settings as the two-stage method Voxel RCNN \cite{deng2021voxel} implemented in OpenPCDet \footnote{https://github.com/open-mmlab/OpenPCDet}. The total number of training epochs is 80. The optimization method employed is the one-cycle Adam \cite{kingma2014adam,smith2019super}, with an initial learning rate of 0.01, weight decay of 0.01, and momentum of 0.9. The IoU thresholds for foreground, background, and box regression in the detection head are set to 0.75, 0.25, and 0.55 (denoted as $\theta_{h}$, $\theta_{l}$, $\theta_{reg}$). All implementations in this paper are based on the OpenPCDet framework.
\begin{table}
	\caption {The MS$^{2}$3D model encoding parameters ($i=0,1,2,3,4$).}
	\label{table1}
	\centering
	{
		\resizebox{\linewidth}{!}{			
			\begin{tabular}{c||c||c} 	
				\hline \hline       				
				Module& Block &	Parameters\\
				\hhline{-||-||-}  	
				\multirow{6}{*}{Multi-Scale Voxelization $\times4$} & FCN$_{C}$ & $10 \longrightarrow C^{v}_{i,1}/2 $\\
				& FCN$_{P}$ &$C^{v}_{i,0} \longrightarrow C^{v}_{i,1}/2$ \\
				&Concatenate&N/A\\
				& FCN &$ C^{v}_{i,1} \longrightarrow C^{v}_{i,2} $\\
				& FCN &$ C^{v}_{i,2} \longrightarrow  C^{v}_{i,2} /2 \longrightarrow  C^{v}_{i,2}/2 \longrightarrow C^{v}_{i,2}$\\
				& $TopK_{i}$ & 40\% $i=0$, 40\% $i=1$, 60\% $i=2$\\
				\hhline{-||-||-} 
				\multirow{6}{*}{3D Encoder $\times4$}& Submanifold Convolution & $3\times3, 1, 2, C^{v}_{i,2} \times 2$\\
				& Sparse Convolution & $3\times3, 1, 1, C^{v}_{i,2} \times 2$\\
				& Sparse Convolution & $3\times3, 1, 1, C^{v}_{i,2} \times 2$\\
				& Submanifold Convolution & $3\times3, 1, stride, C^{v}_{i,2} \times 2$\\
				& Sparse Convolution & $3\times3, 1, 1, C^{v}_{i,2} \times 2$\\
				& Sparse Convolution & $3\times3, 1, 1, C^{v}_{i,2} \times 2$\\
				\hhline{-||-||-} 
				\multirow{3}{*}{2D Encoder $\times4$} & Base Block & $96,i=0,1$ or $128,i=1,2$ \\
				& Base Block & 128 \\
				& Base Block & 256 \\
				\hline \hline    
	\end{tabular}} }  	
\end{table}
\begin{table*}
	\captionsetup{justification=raggedright,singlelinecheck=false}
	\caption {Results of 3D detection benchmarks on the KITTI test set. ‘V’ denotes voxel-based methods. ‘P’ denotes point-based methods. ‘P+V’ denotes mixed point-based and voxel-based methods. ‘L’ denotes the usage of only LiDAR data as input. ‘L+I’ denotes using both LiDAR and image data as input. Bold indicates the best score for all methods, and underlined indicates the best score for the voxel-based method.}
	\label{table2}
	\centering
	\footnotesize	
		\begin{tabular}{ccc||ccccccccc}  	
		\hline \hline    				
		\multirow{2}{*}{Method} & 
		\multirow{2}{*}{Type} & 
		\multirow{2}{*}{Sens.} &    	
		\multicolumn{3}{c}{Car $AP_{3D}$(\%)} & 
		\multicolumn{3}{c}{Pedestrian $AP_{3D}$(\%)} &
		\multicolumn{3}{c}{Cyclist $AP_{3D}$(\%)} \\ 	  
		\cline{4-12}     			
		&  &  & Easy & Mod. & Hard & Easy & Mod. & Hard & Easy & Mod. & Hard \\ 
		\hhline{---||---------}     					
		MV3D\cite{chen2017multi}& - & \textbf{$L+I$}&74.97&63.63&54.00&-&-&-&-&-&-\\
		F-ConvNet\cite{wang2019frustum}& - & \textbf{$L+I$}&87.36&76.39&66.69&-&-&-&-&-&-\\ 
		IPOD\cite{yang2018ipod}& - & \textbf{$L+I$}&71.40&53.46&48.34&-&-&-&-&-&-\\ 
		EPNet\cite{huang2020epnet}& - & \textbf{$L+I$}&59.81&79.28&74.50&-&-&-&-&-&-\\ 
		ContFuse\cite{liang2018deep}& - & \textbf{$L+I$}&83.68&68.78&61.67&-&-&-&-&-&-\\ 
		CT3D\cite{sheng2021improving}& - & \textbf{$L+I$}&87.83&\textbf{81.77}&77.16&-&-&-&-&-&-\\ 
		AVOD\cite{ku2018joint}& - & \textbf{$L+I$}&83.07&71.76&63.73&50.46&42.27&39.04&63.76&50.55&44.93\\ 
		F-PointNet\cite{qi2018frustum}& - & \textbf{$L+I$} &82.19&69.76&60.59&50.53&42.15&38.08&72.27&56.12&49.01\\ 
		\hhline{---||---------}  
		3DSSD\cite{yang20203dssd}&\textbf{$P$}& \textbf{$L$}&88.36&79.57&74.55&-&-&-&-&-&-\\ 
		IA-SSD\cite{zhang2022not}&\textbf{$P$}& \textbf{$L$}&88.34&80.13&75.04&46.51&39.03&35.60&78.35&61.94&55.70\\
		STD\cite{yang2019std}&\textbf{$P$}& \textbf{$L$}&87.95&79.71&75.09&53.29&42.49&38.55&78.69&61.59&55.30\\ 
		Point-GNN\cite{shi2020point}&\textbf{$P$}& \textbf{$L$}&88.33&79.47&74.55&54.64&44.27&40.23&\textbf{82.48}&64.10&56.90\\  
		Point RCNN\cite{shi2019pointrcnn}&\textbf{$P$}& \textbf{$L$}&85.94&75.76&68.32&49.43&41.78&38.63&78.58&62.73&57.74\\ 
		\hhline{---||---------} 
		BADet\cite{qian2022badet}&\textbf{$P+V$}& \textbf{$L$}&89.28&81.61&76.58&-&-&-&-&-&-\\  
		HVPR\cite{noh2021hvpr}&\textbf{$P+V$}& \textbf{$L$}&86.38&77.92&73.04&53.47&43.96&40.64&-&-&-\\ 
		PV-RCNN\cite{shi2020pv}&\textbf{$P+V$}& \textbf{$L$}&92.25&81.43&76.82&52.17&43.29&40.29&78.60&63.71&57.65\\
		\hhline{---||---------}   
		SECOND\cite{yan2018second}&\textbf{$V$}& \textbf{$L$}&83.34&72.55&65.82&-&-&-&-&-&-\\
		Voxel RCNN\cite{deng2021voxel}&\textbf{$V$}& \textbf{$L$}&90.90&81.62&77.06&-&-&-&-&-&-\\
		SA-SSD\cite{he2020structure}&\textbf{$V$}& \textbf{$L$}&88.75&79.79&74.16&-&-&-&-&-&-\\   
		PartA$^{2}$\cite{shi2020points}&\textbf{$V$}& \textbf{$L$}&85.94&77.86&72.00&\underline{\textbf{54.49}}&44.50&\underline{\textbf{42.36}}&78.58&62.73&57.74\\  
		PointPillars\cite{lang2019pointpillars}&\textbf{$V$}& \textbf{$L$}&82.58&74.31&68.99&51.45&41.92&38.89&77.10&58.65&51.92\\ 
		H$^{2}$3D RCNN\cite{deng2021multi}&\textbf{$V$}& \textbf{$L$}&\underline{\textbf{92.43}}&81.55&\underline{\textbf{77.22}}&52.75&45.26&41.65&78.67&62.74&55.78\\  
		ASCNet\cite{tong2022ascnet}&\textbf{$V$}& \textbf{$L$}&88.48&\underline{81.67}&76.93&42.00&35.76&33.69&78.41&\underline{\textbf{65.10}}&\underline{\textbf{57.87}}\\ 
		SIEV-Net\cite{9773165}&\textbf{$V$}&\textbf{$L$}&85.21&76.18&70.60&54.00&\underline{\textbf{44.80}}&41.11&78.75&59.99&52.37\\ 
		SMS-Net\cite{liu2022sms}&\textbf{$V$}& \textbf{$L$}&87.01&76.21&70.45&53.46&44.76&41.35&75.35&60.23&53.37\\ 
		\hhline{---||---------} 
		Ours&\textbf{$V$}& \textbf{$L$}&88.05&79.64&74.93&51.83&42.73&39.49&\underline{80.48}&64.81&57.21\\  
		\hline \hline     			
	\end{tabular} 
\end{table*}
\begin{figure*}
	\centering
	\includegraphics[width=7.25in]{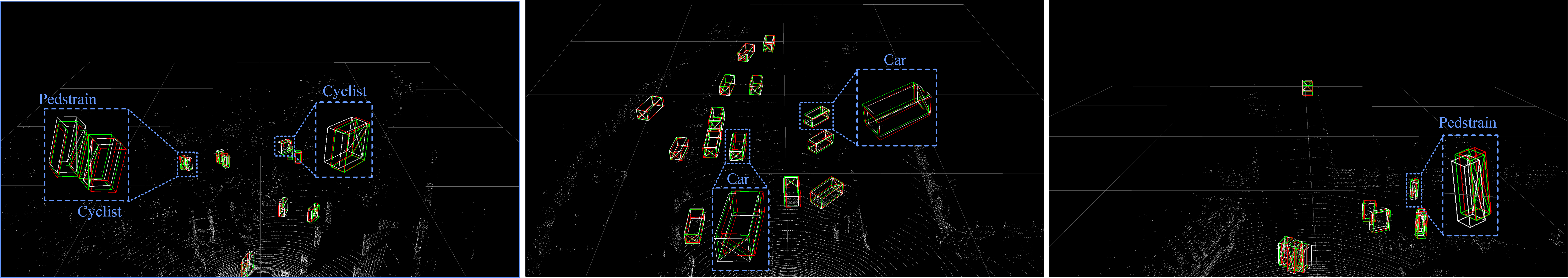}
	\caption{Visualization of detection results of our method and Voxel RCNN on the KITTI dataset. The green boxes represent ground truth bounding boxes, the red boxes represent the detection results of our method, and the white boxes represent the detection results of Voxel RCNN.}
	\label{fig_8}
\end{figure*}
\begin{figure*}
	\centering
	\includegraphics[width=7.25in]{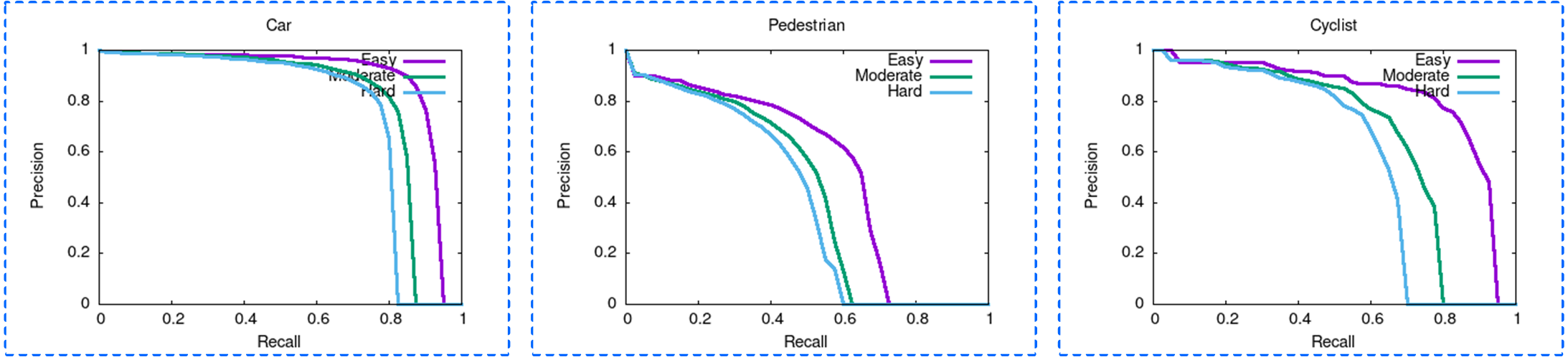}
	\caption{Visualization of the Precision-Recall (P-R) curves of ‘Car’, ‘Pedestrian’, and ‘Cyclist’ 3D detection results on the KITTI test dataset. The purple, green, and blue curves represent the curves under easy, moderate, and hard conditions, respectively; the horizontal axis of the P-R curve represents recall, while the vertical axis represents precision.}
	\label{fig_9}
\end{figure*}
\begin{table}[h]
	\caption {Results of 3D detection benchmarks on the KITTI validation set (3,712 samples for training and 3,769 samples for validation). The bolded portions represent the best results. The detection speed and memory usage are done on a device with a GTX1060.}
	\label{table3}
	\LARGE
	\centering
	\resizebox{\linewidth}{!}{      				
		\begin{tabular}{cc||cccccccc}
			\hline \hline    				
			\multirow{2}{*}{Method} & 
			\multirow{2}{*}{Type} &   	
			\multicolumn{3}{c}{Hard} & 
			\multicolumn{3}{c}{Mod $\setminus$ Hard (Drop 40\%)}&
			\multirow{2}{*}{FPS}&\multirow{2}{*}{Mem.}\\ 
			\cline{3-8}   	  		
			&&Car&Ped.&Cyc.&Car&Ped.&Cyc.\\
			\hhline{--||--------}   				
			PV-RCNN\cite{shi2020pv}&P+V&77.86&45.02&60.96&77.35$\setminus$75.55&47.90$\setminus$43.48&58.70$\setminus$54.95&3.8&1439Mib\\
			Point RCNN\cite{shi2019pointrcnn}&P&76.10&51.76&65.49&72.13$\setminus$68.53&51.95$\setminus$44.89&57.86$\setminus$54.54&5&1093Mib\\
			PartA$^{2}$\cite{shi2020points}&V&78.50&44.53&\textbf{65.67}&\textbf{77.97}$\setminus$73.04&48.89$\setminus$42.04&60.68$\setminus$57.08&3&1021Mib\\
			Voxel RCNN\cite{deng2021voxel}&V&\textbf{79.31}&48.15&64.03&77.48$\setminus$\textbf{75.98}&49.30$\setminus$43.96&61.62$\setminus$56.74&9&991Mib\\
			PointPillars\cite{lang2019pointpillars}&V&72.87&42.02&57.16&71.13$\setminus$68.97&43.33$\setminus$38.89&48.36$\setminus$43.10&\textbf{14}&\textbf{931Mib}\\
			SECOND\cite{yan2018second}&V&75.90&42.55&61.71&76.01$\setminus$73.11&44.33$\setminus$40.02&53.36$\setminus$51.10&11&889Mib\\
			H$^{2}$3D RCNN\cite{deng2021multi}&V&78.63&\textbf{52.56}&56.83&77.76$\setminus$75.41&51.01$\setminus$45.06&50.25$\setminus$49.67&12&941Mib\\
			Ours&V&78.43&50.33&63.93&77.46$\setminus$74.48&\textbf{52.11}$\setminus$\textbf{47.08}&\textbf{62.47}$\setminus$\textbf{57.14}&8&1229Mib\\
			\hline \hline      			
		\end{tabular}} 
\end{table}
\subsection{Result on KITTI Dataset}
We evaluate the 3D Average Precision (3D AP) for three categories: Car, Cyclist, and Pedestrian, using IoU thresholds of 0.7 for Car and 0.5 for Cyclist and Pedestrian. The detection results obtained from our method are submitted to the KITTI test dataset and are presented in \hyperref[table2]{Table 2}. We visualize our detection results on the KITTI dataset in \hyperref[fig_7]{Figure 7}. In \hyperref[fig_8]{Figure 8}, we have shown the results of our method and Voxel RCNN.
\begin{table*}[h]
	\caption {Results on the ONCE validation set.}
	\label{table4}
	\centering
	\LARGE
	\resizebox{\linewidth}{!}{      				
		\begin{tabular}{cc||ccccccccccccc}  	
			\hline \hline     				
			\multirow{2}{*}{Method} & 
			\multirow{2}{*}{Type} &   	
			\multicolumn{4}{c}{Vehicle} & 
			\multicolumn{4}{c}{Pedestrian} &
			\multicolumn{4}{c}{Cyclist} &
			\multirow{2}{*}{mAP}\\ 	  
			\cline{3-14}     			
			&&$overall$&$0-30m$&$30-50m$&$>50m$&$overall$&$0-30m$&$30-50m$&$>50m$&$overall$&$0-30m$&$30-50m$&$>50m$&\\ 
			\hhline{--||-------------}     				
			PV-RCNN\cite{shi2020pv}&\textbf{$P+V$}&77.77&89.39&72.55&58.64&23.50&25.61&22.84&17.27&59.37&71.66&52.58&36.17&53.55\\
			Point RCNN\cite{shi2019pointrcnn}&\textbf{$P$}&52.09&74.45&40.89&16.81&4.28&6.17&2.40&0.91&29.84&46.03&20.94&5.46&28.74\\
			PointPillars\cite{lang2019pointpillars}&\textbf{$V$}&68.57&80.86&62.07&47.04&17.63&19.47&15.15&10.23&46.81&58.33&40.32&25.86&44.34\\
			CenterPoints\cite{yin2021center}&\textbf{$V$}&66.79&80.10&59.55&43.39&49.90&56.24&42.61&26.27&63.45&74.28&57.94&41.48&60.05\\
			SECOND\cite{yan2018second}&\textbf{$V$}&71.19&84.04&63.02&47.25&26.44&29.33&24.05&18.05&58.04&69.69&52.43&34.61&51.86\\
			Ours&\textbf{$V$}&74.08&86.09&67.05&51.27&26.84&32.33&20.67&12.80&56.12&69.94&47.31&28.64&52.35\\
			\hline \hline      			
	\end{tabular}} 
\end{table*}

In \hyperref[table2]{Table 2}, we categorize 3D object detection methods based on point clouds into two groups: LiDAR-based methods and fusion methods that combine image and LiDAR data. Furthermore, LiDAR-based methods are subdivided into three types: (1) point-based, (2) voxel-based (considering that both voxel-based and projection-based methods transform the point clouds into a regular grid-like representation, we categorize both types as voxel-based in \hyperref[table2]{Table 2}), and (3) point \& voxel-based. The proposed MS$^{2}$3D in this paper belongs to the voxel-based category, utilizing only point clouds as input (it's essential to note that we employ a multi-class joint training method, where a single model is trained to detect objects from multiple classes). The final detection results are then submitted to the KITTI dataset. Additionally, for a more comprehensive evaluation of our method, in \hyperref[fig_9]{Figure 9}), we present the Precision-Recall (P-R) curves of ‘Car’, ‘Pedestrian’, and ‘Cyclist’ 3D detection results on the KITTI test dataset.

As depicted in \hyperref[table2]{Table 2}, voxel-based methods generally exhibit higher detection accuracy for the ‘Car‘ category but lower accuracy for smaller objects such as ‘Cyclist’ and ‘Pedestrian’, which has fewer reflective points. In contrast, our method performs well on ‘Cyclist’ and ‘Pedestrian’. Compared to PartA$^{2}$, which exhibits the best performance on ‘Pedestrian’, our method has a difference of 2-3\%. However, for the ‘Cyclist’, our detection accuracy is higher by 2-3\%, and for the ‘Car’, our detection accuracy is also higher by 2-3\%. Additionally, compared to H$^{2}$3D RCNN, our detection accuracy on ‘Cyclist’ is also higher by 2-3\%. In addition, as shown in \hyperref[fig_8]{Figure 8}, compared to the classical two-stage method Voxel RCNN, our method also performs better on ‘Pedestrian’ and ‘Cyclist’. 

This can be attributed to the following factors: (1) Loss of fine-grained features: Voxel-based methods tend to increase the receptive field of feature points through convolutional downsampling, capturing distant semantic features with larger receptive fields but potentially missing fine local features. (2) Foreground Point Retention: In autonomous driving scenarios, the majority of points constitute background, while only a few points are foreground and relevant for detection. Voxel-based downsampling may result in discarding some foreground points. (3)Sparse 3D feature layer: LiDAR-generated point clouds are sparse, especially for distant, small-sized, and low-reflective objects. Constructing a 3D feature layer through voxelization exacerbates the sparsity of initially sparse feature points, making it challenging to retain geometric features. (4) Object Centroid and Feature Aggregation: LiDAR point clouds often capture only the surfaces of objects, with the centroid of an object frequently distant from feature points.
\begin{figure}
	\centering
	\includegraphics[width=3.5in]{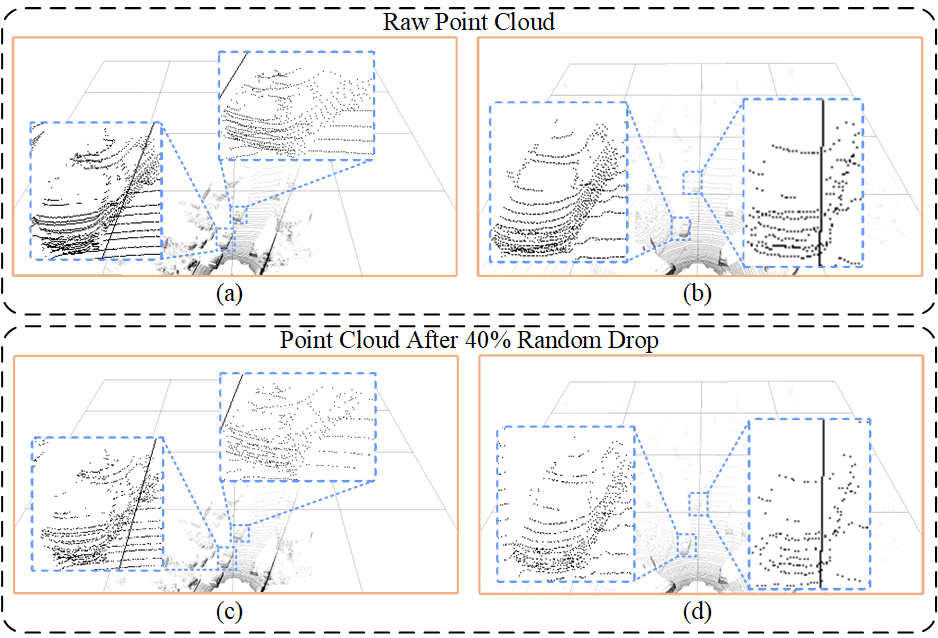}
	\caption{Visualization of a randomly dropped 40\% point cloud.Where (a),(b) denote the raw point cloud scene, and (c),(d) denote the scene after randomly dropping 40\% of the point cloud.}
	\label{fig_10}
\end{figure}

In our method: (1) Multi-scale Voxelization is utilized to generate both long-range semantic features and fine-grained local features; (2) In the encoder process of Multi-Scale Voxelization, distance weights ($W_{d}$) are computed to preserve foreground points closer to object's centroid, with weighted Mean operations further emphasizing foreground feature; (3) Various levels of sparse voxel features contribute to a denser 3D feature layer with diverse semantic features; (4) Motivated by VoteNet \cite{zhang2022not}, we employ Center Vote to predict the offsets of deep-level feature points from the object's centroid, allowing them to aggregate features more effectively for subsequent stages. Additionally, compared to Voxel-based methods, Point-based methods typically mitigate feature loss. Our method also exhibits promising competitiveness compared to Point-based methods and the Point \& Voxel-based PV-RCNN.
\begin{figure*}
	\centering
	\includegraphics[width=7.25in]{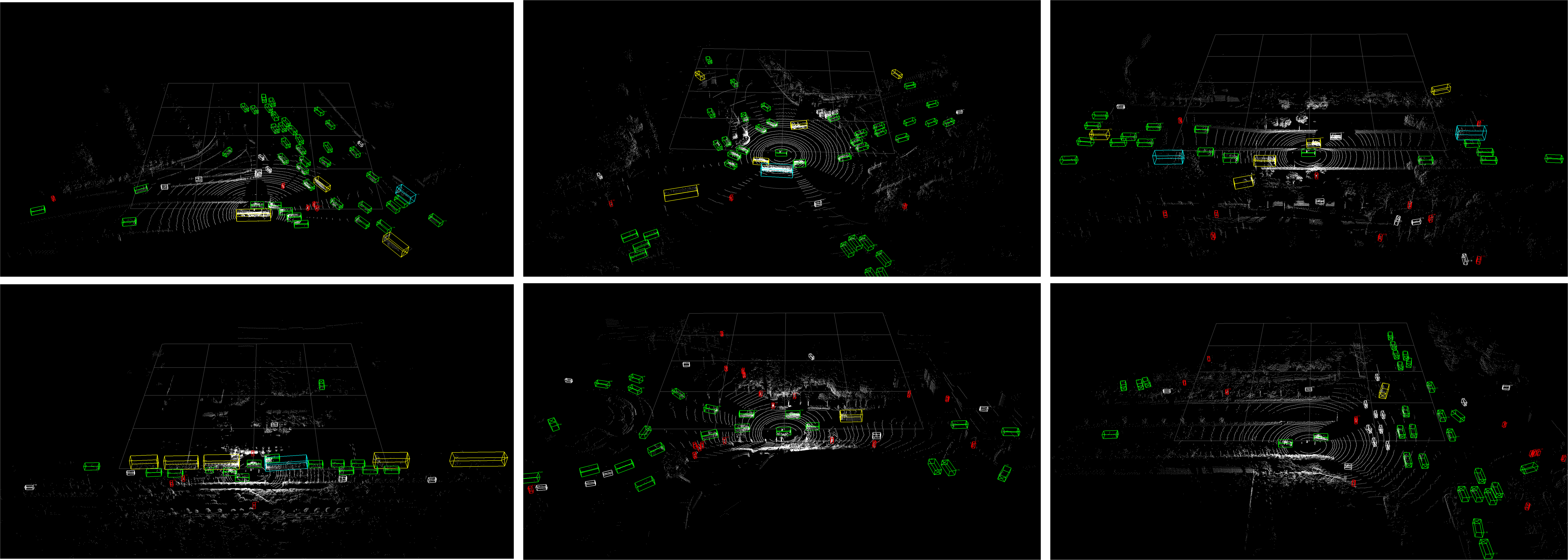}
	\caption{Visualization of the detection results on the ONCE dataset. The yellow boxes represent ‘Truck’, the green boxes represent ‘Car’, the blue boxes represent ‘Bus’, the red boxes represent ‘Pedestrian’, and the white boxes represent ‘Cyclist’.}
	\label{fig_11}
\end{figure*}

Current study on 3D object detection methods mostly relies on high-line Lidar (for example, the KITTI dataset is created using a 64-line Lidar). While point clouds generated by high-line Lidar are sparse, those generated by low-line Lidar are even sparser in comparison. This sparsity limits their ability to accurately describe the geometric features of objects. Therefore, when using low-line Lidar, the performance of most 3D object detection methods is poor. However, due to cost constraints, it is necessary to research the generalization of 3D object detection to sparse point clouds generated by low-line Lidar. To simulate the point cloud data generated by a low-line LiDAR, we randomly dropped 40\% of the point clouds in the validation subset (3,769 samples) of the KITTI training dataset. Comparing \hyperref[fig_10]{Figure 10 (a) (b)} with \hyperref[fig_10]{(c) (d)}, it can be observed that after dropping 40\% of the point clouds, objects lose most of their detailed features. Especially for distant objects, only basic geometric features are retained. This poses significant challenges to the detection tasks. To validate the generalization of our method, we trained our method and other 3D object detection methods on the training subset (3,712 samples) of the KITTI training dataset (without dropping any point clouds) and conducted comparative experiments on the validation subset with 40\% of the point clouds randomly dropped. The final experimental results are shown in \hyperref[table3]{Table 3}.

From \hyperref[table3]{Table 3}, it is evident that under the Hard conditions, our method demonstrates commendable detection performance for pedestrians and cyclists. While \hyperref[table2]{Table 2} highlights a gap between our method and Voxel RCNN as well as H$^{2}$3D RCNN, \hyperref[table3]{Table 3} shows that our method maintains commendable detection accuracy even with a relatively sparse point cloud environment. This is attributed to the Semantic Feature Aggregation module proposed in this paper, which leverages semantic feature points at various levels to construct a more compact 3D feature layer. MS$^{2}$3D is capable of generating a relatively compact 3D feature layer even with a sparser point cloud environment. This transformation effectively increases the point cloud's density, thereby ensuring robust detection accuracy in environments with sparser point clouds.

To validate the real-time performance and memory usage of our method, we conduct tests on a device equipped with GTX1060 (comparable to the computational power of automotive autonomous driving device), and the results are shown in \hyperref[table3]{Table 3}. On the device equipped with GTX1060, our method achieves a detection speed of 8Hz (i.e., detecting 8 frames of point clouds per second, which surpasses most two-stage methods). Compared to PointPillars, SECOND (one-stage methods), and H$^{2}$3D RCNN (two-stage method), our method has a certain gap in detection speed. However, as shown in \hyperref[table3]{Table 3}, our method maintains an absolute advantage in terms of detection accuracy in sparse scenes. In summary, while meeting the requirements for real-time detection in autonomous driving scenarios, our method also performs excellently in terms of detection accuracy, especially in sparse scenes.
\subsection{Result on ONCE Dataset}
To further assess the performance of MS$^{2}$3D in more intricate scenarios, we conduct detections using MS$^{2}$3D on the validation set of the ONCE dataset. As demonstrated in \hyperref[table4]{Table 4}, MS$^{2}$3D remains competitive compared to other voxel-based methods. This reaffirms that the proposed method can achieve significant performance even in complex and large-scale LiDAR scenes. Additionally, in \hyperref[fig_11]{Figure 11}, we present visualizations of our detection results on the ONCE dataset.
\subsection{Ablation Experiments}
In this section, we conduct ablation experiments to evaluate the three Center Vote schemes mentioned in Semantic Feature Aggregation. All experiments are carried out on the same hardware with consistent parameter settings for training. The ultimate results are evaluated on the KITTI validation set.
\begin{figure}
	\centering
	\includegraphics[width=3.5in]{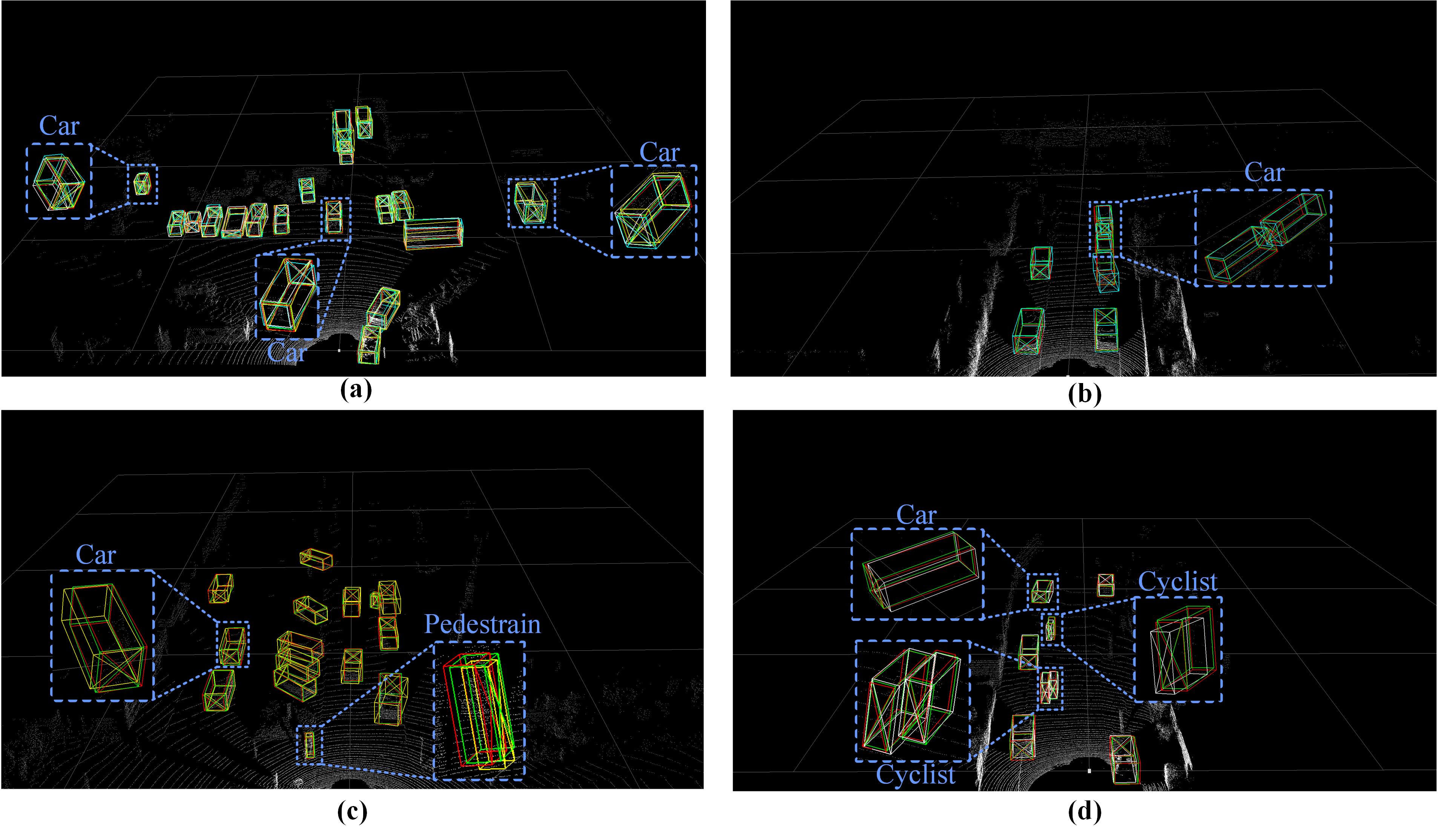}
	\caption{Visualization of the detection results of methods in \hyperref[table5]{Table 5}. The green boxes represent ground truth bounding boxes, the white boxes represent the detection results of method (3), the blue boxes represent the detection results of method (4), the yellow boxes represent the detection results of method (5), and the red boxes represent the detection results of method (6).}
	\label{fig_12}
\end{figure}

The results of the experiment are shown in \hyperref[table5]{Table 5}. In this table, ‘Centerness’ refers to the use of $L_{ctr}$ for supervising $W_{d}$; ‘MV-Vote’ indicates Center Vote applied to the last layer of Multi-Scale Voxelization; ‘All-Vote’ indicates Center Vote applied to all feature points in the 3D feature layer; ‘$V_{i}$-Vote’ indicates Center Vote applied to $V_{i}$ in the 3D feature layer; ‘$F_{i}$-Vote’ signifies Center Vote applied to $F_{i}$ in the 3D feature layer.

\textbf{Ablation Experiments with Distance Weights:} In Multi-Scale Voxelization, we propose a method of using distance weights ($W_{d}$) to select feature points more conducive to object detection and emphasize their features. To analyze this, we conduct experiments with and without supervised signals provided to $W_{d}$ during training. Method (1) in \hyperref[table5]{Table 5} does not provide supervised signals to $W_{d}$. Method (2) involves supervising $W_{d}$ through $L_{ctr}$. Comparing the results of methods (1) and (2) in \hyperref[table5]{Table 5}, we observe that method (2) yields improved detection accuracy for both ‘Car’ and ‘Cyclist’. This is because supervised signals guide $W_{d}$ to assign greater weights to feature points closer to instance centers, and during Multi-Scale Voxelization's sampling process, it selects foreground points as much as possible.

\textbf{Ablation Experiments with Center Vote in Multi-Scale Voxelization:} In Multi-Scale Voxelization, we propose performing Center Vote on the last layer ($V_{4}$) of Multi-Scale Voxelization. We conduct experiments to analyze the effects of this approach. Method (2) in \hyperref[table5]{Table 5} indicates no Center Vote, while method (3) performs Center Vote on $V_{4}$. By comparing the experimental results, we observe that method (3) shows a slight improvement over method (2). This enhancement is attributed to the rich semantic features contained in $V_{4}$, combined with its substantial receptive field. Center Vote on $V_{4}$ enables it to shift closer to the object's centroid, facilitating subsequent RPN network-generated region proposals.

\textbf{Ablation Experiments with Semantic Feature Aggregation:} In Semantic Feature Aggregation, we introduce three Center Vote schemes for the 3D feature layer. Method (3) doesn't involve Center Vote in the 3D feature layer; method (4) applies Center Vote to all feature points in the 3D feature layer; method (5) conducts Center Vote on feature points ($V_{i}$) from Multi-Scale Voxelization within the 3D feature layer; method (6) performs Center Vote on feature points ($F_{i}$) from the 3D Encoder within the 3D feature layer.

When comparing method (3) and (6), it's clear that method (6) demonstrates improvements for both cars and pedestrians, particularly notable with a 5\% enhancement for pedestrians. Analysis attributes this improvement to LiDAR capturing fewer measurements for targets with fewer reflective points, such as pedestrians. Effective feature aggregation is challenging if the region proposals generated in the first stage lack precision, possibly resulting in the aggregation of irrelevant foreground and background points, or failing to aggregate all the necessary foreground points. Our method proposed can avoid the impact of poor-quality region proposals on feature aggregation, achieving better performance in subsequent detection tasks.

Upon comparing the experimental results in \hyperref[table5]{Table 5}, it is evident that method (6) exhibits a substantial improvement over method (3), whereas methods (4) and (5) present inferior detection results compared to method (3). 

In \hyperref[fig_12]{Figure 12}, we have further presented the visualization results. As shown in \hyperref[fig_12]{Figure 12 (b)}, it can be observed that even for objects at close distances with a significant number of reflection points, the detection results of method (4) exhibit substantial errors. As depicted in \hyperref[fig_5]{Figure 5}, applying Center Vote to all feature points in method (4) severely distorts the object geometry. Specifically, in the z-axis direction, the 3D feature layer constructed by method (4) shifts all z-axis points onto the plane of the object's centroid, significantly compromising geometric feature in the z-axis direction. 
\begin{table}
	\caption {Results of ablation experiments. The evaluation is conducted on the KITTI validation set, using 3D AP with recalls@40 under the Mod. condition setting as the accuracy metric. The bolded portions represent the best results.}
	\label{table5}
	\centering
	\LARGE
	\resizebox{\linewidth}{!}{      				
		\begin{tabular}{cccccc||ccc} 	
			\hline \hline 				
			\multirow{2}{*}{Method} & 
			\multirow{2}{*}{Centerness} &   	
			\multirow{2}{*}{MV-Vote} & 
			\multirow{2}{*}{All-Vote} & 
			\multirow{2}{*}{$V_{i}$-Vote} & 
			\multirow{2}{*}{$F_{i}$-Vote} & 
			Car Mod. & Ped. Mod.& Cyc. Mod.\\ 	  		
			&&&&&&$(IoU=0.7)$&$(IoU=0.5)$&$(IoU=0.5)$\\
			\hhline{------||---}     				
			(1)&-&-&-&-&-&78.69&\textbf{58.02}&63.09\\
			(2)&$\surd $&-&-&-&-&82.34&57.49&64.00\\
			(3)&$\surd $&$\surd $&-&-&-&82.53&58.13&64.34\\
			(4)&$\surd $&$\surd $&$\surd $&-&-&79.96&56.93&62.36\\
			(5)&$\surd $&$\surd $&-&$\surd $&-&81.36&55.42&63.79\\
			(6)&$\surd $&$\surd $&-&-&$\surd $&\textbf{82.70}&57.87&\textbf{69.84}\\
			\hline \hline     			
	\end{tabular}} 
\end{table}

Method (5) demonstrates improved detection accuracy compared to method (4). As shown in \hyperref[fig_12]{Figure 12 (a), (b), (c)}, method (5) has smaller detection errors compared to method (4). However, when compared to method (6), method (5) still shows significant errors for objects with fewer reflection points, such as pedestrians. As shown in \hyperref[fig_5]{Figure 5}, although method (5) avoids having all feature points clustered near the object's centroid, it still struggles to accurately describe its geometric structure. Additionally, for objects with fewer reflection points like pedestrians, there are fewer deep-level feature points in the 3D feature layer. We hope that these points can be aggregated as much as possible for subsequent detection tasks. When the quality of region proposals generated is poor, method (5) may lead to the loss of most deep-level feature points, affecting subsequent detection tasks.

As shown in \hyperref[table5]{Table 5} and \hyperref[fig_12]{Figure 12}, method (6) achieves the best detection results. From \hyperref[fig_12]{Figure 12 (d)}, it can be observed that, compared to method (3), method (6) performs well for objects with fewer reflection points. Method (6) only applies Center Vote to feature points $F_{i}$, guiding the deep-level feature points $F_{i}$ with rich semantic features to a position near the object's centroid. This avoids the impact of poor-quality region proposals on feature aggregation. Additionally, retaining shallow-level feature points on the object's surface also avoids the influence on the geometric features of objects, as observed in Methods (4) and (5).
\begin{table}	
	\caption {Results of ablation experiments on distance-weighted sampling. The evaluation is conducted on the KITTI validation set, using 3D AP with recalls@40 under the Mod. condition setting as the accuracy metric. The bolded portions represent the best results. $TopK_{i}$ represents the sampling rates used in the i-th downsampling of Multi-Scale Voxelization.}	\label{table6}	
	\centering	
	\LARGE	
	\resizebox{\linewidth}{!}{      						\begin{tabular}{c||ccc||ccc} 				
			\hline \hline     							
			\multirow{2}{*}{Method} & 			
			\multirow{2}{*}{$TopK_{1}$} &   				
			\multirow{2}{*}{$TopK_{2}$} & 			
			\multirow{2}{*}{$TopK_{3}$} & 			
			Car Mod. & Ped. Mod.& Cyc. Mod.\\ 	  					&&&&$(IoU=0.7)$&$(IoU=0.5)$&$(IoU=0.5)$\\			
			\hline 			
			(1)&20\%&20\%&20\%&79.51&54.99&66.04\\			(2)&40\%&40\%&40\%&80.95&56.08&68.35\\			
			(3)&40\%&40\%&60\%&\textbf{82.70}&\textbf{57.87}&\textbf{69.84}\\			(4)&60\%&60\%&60\%&81.77&57.62&68.73\\			(5)&80\%&80\%&80\%&82.44&57.09&68.52\\      							
			\hline \hline     				
		\end{tabular}} 
	\end{table}
In \hyperref[table6]{Table 6}, we conduct further experimental validation on the sampling rates used in downsampling during Multi-Scale Voxelization. $TopK_{i}$ in \hyperref[table6]{Table 6} represents the sampling rate used in the i-th downsampling step (selecting the top k percent of feature points based on the distance-weighted Wd for subsequent detection tasks; method (3) denotes the sampling rate used in this paper). From \hyperref[table6]{Table 6}, it can be observed that when the sampling rates are $[40\%, 40\%, 60\%]$, the detection results perform the best. The main reasons maybe as follows: (1) When the sampling rate is lower than the optimal value (methods (1) and (2)), some foreground feature points are lost as downsampling progresses, which is detrimental to the subsequent description of object geometric features. (2) When the sampling rate is higher than the optimal value (methods (1) and (2)), increasing the sampling rate further will not introduce more foreground points due to constraints on the number of foreground points. Furthermore, it may introduce some background points, which will have a certain impact on the final detection results.

\section{Conclusion}
The voxel-based two-stage 3D object detection method completes the final detection task based on the results of the first stage and has achieved satisfactory performance. However, in autonomous driving scenarios, the sparsity and hollowness of point clouds still pose challenges for voxel-based two-stage methods. As for the sparsity and hollowness of point clouds, this paper proposes a two-stage 3D object detection framework called MS$^{2}$3D.

In response to the sparsity of the point clouds, we propose a new method, which use voxel feature points from multiple branches to constructing the 3D feature layer and samples feature points based on distance weight. By employing voxel feature points from different branches, we successfully build a relatively compact 3D feature layer with rich semantic features. The distance-weighted sampling method helps to preserve foreground feature points as much as possible. To validate the effectiveness of our method, we visualize the 3D feature layer, proving that our method adequately describes the geometric features of objects. In the experiments on the KITTI dataset, our method has shown good performance in detecting cyclists and pedestrians with fewer reflective points (the 3D detection accuracy for cyclists under Mod. conditions is 64.81\%, and for pedestrians, it is 42.73\%). In addition, to validate the performance of our method in sparse scenes, we randomly dropped 40\% of the point clouds on the KITTI dataset to simulate sparse scenarios. The final results also demonstrate that our method performs well in sparse point cloud scenes (3D detection mAP under Mod. conditions is 64.01\%).

In response to the hollowness of point clouds, we propose a new method, which predict the offsets between deep-level feature points and the object's centroid, making them as close as possible to the object's centroid. While for shallow-level feature points , we retain them on the object's surface. We visualize the 3D feature layer and demonstrate that moving the deep-level feature points to the object's centroid can aggregate rich semantic features while preserving the geometric features of the object. The results of ablation experiments and visualizations of detection results in the ablation experiment validate the feasibility of our method. The 3D detection mAP under Mod. conditions is as follows: only offsetting the deep-level feature points towards the object's centroid: 70.14\%; only offsetting the shallow-level feature points towards the object's centroid: 66.86\%; only offsetting all feature points towards the object's centroid: 66.42\%; no offset: 68.33\%).

In autonomous driving scenarios, there is a need to tackle more complex scenarios, particularly in detecting distant and small objects. The inherent imaging nature of LiDAR result in fewer reflection points for objects at distant and small objects. Although our method performs well for objects with fewer reflection points, the imaging natures of LiDAR make it challenging to describe the geometric features of them using LiDAR point clouds. This poses a difficulty in further improving our method. In the future, we plan to address this issue by utilizing cameras to generate pseudo point clouds. This method aims to compensate for the insufficient description of geometric features in LiDAR point clouds for objects with fewer reflection points.
\section{Acknowledgments}
This research was funded by the Provincial Natural Science Foundation of Zhejiang, grant number LY21F010013.



\bibliographystyle{elsarticle-num-names} 
\bibliography{reference.bib}


%
%
%
\end{document}